%% file: main.tex
\documentclass[10pt,twocolumn,letterpaper]{article}

\usepackage[pagenumbers]{cvpr} 
\usepackage{algorithmic}
\usepackage{algorithm}
\usepackage{adjustbox}
\usepackage{multirow}
\usepackage{colortbl}
\usepackage{adjustbox}
\usepackage{tabularx}
\usepackage{marvosym}    

\input{preamble}
\definecolor{cvprblue}{rgb}{0.21,0.49,0.74}
\usepackage[pagebackref,breaklinks,colorlinks,allcolors=cvprblue]{hyperref}

\def\paperID{} 
\def\confName{CVPR}
\def\confYear{2026}

\title{MUSE: Harnessing Precise and Diverse Semantics for Few-Shot Whole Slide Image Classification }

\author{Jiahao Xu\textsuperscript{1}, Sheng Huang\textsuperscript{1}\thanks{Corresponding author}, Xin Zhang\textsuperscript{1}, Zhixiong Nan\textsuperscript{2}, Jiajun Dong\textsuperscript{1} and Nankun Mu\textsuperscript{2}\\ 
\textsuperscript{1}School of Big Data \& Software Engineering, Chongqing University, \\ \textsuperscript{2}School of Computer Science \& Technology, Chongqing University\\
{\tt\small \{xujiahao,zhangxin,dongjiajun\}@stu.cqu.edu.cn, \{huangsheng,nanxz,nankun.mu\}@cqu.edu.cn}
}

\begin{document}
\maketitle
\input{sec/0_abstract}    
\input{sec/1_intro_HS}

\input{sec/2_relatedHS}
\input{sec/3_method_HS}
\input{sec/4_experiment}

\input{sec/5_conclusion}

 \bibliographystyle{ieeenat_fullname}
 \bibliography{main}

\input{sec/X_suppl}



\end{document}

%% file: sec/0_abstract.tex
\begin{abstract}
In computational pathology, few-shot whole slide image classification is primarily driven by the extreme scarcity of expert-labeled slides. Recent vision-language methods incorporate textual semantics generated by large language models, but treat these descriptions as static class-level priors that are shared across all samples and lack sample-wise refinement. This limits both the diversity and precision of visual-semantic alignment, hindering generalization under limited supervision. To overcome this, we propose the stochastic MUlti-view Semantic Enhancement (MUSE), a framework that first refines semantic precision via sample-wise adaptation and then enhances semantic richness through retrieval-augmented multi-view generation. Specifically, MUSE introduces Sample-wise Fine-grained Semantic Enhancement (SFSE), which yields a fine-grained semantic prior for each sample through MoE-based adaptive visual-semantic interaction. Guided by this prior, Stochastic Multi-view Model Optimization (SMMO) constructs an LLM-generated knowledge base of diverse pathological descriptions per class, then retrieves and stochastically integrates multiple matched textual views during training. These dynamically selected texts serve as enriched semantic supervisions to stochastically optimize the vision-language model, promoting robustness and mitigating overfitting. Experiments on three benchmark WSI datasets show that MUSE consistently outperforms existing vision-language baselines in few-shot settings, demonstrating that effective few-shot pathology learning requires not only richer semantic sources but also their active and sample-aware semantic optimization. Our code is available at: \url{https://github.com/JiahaoXu-god/CVPR2026_MUSE}.

\end{abstract} 

%% file: sec/1_intro_HS.tex
\section{Introduction}
\label{sec:intro}

Computational pathology (CPath)~\cite{cpath1,cpath2,cpath3}, a key branch of digital pathology~\cite{dpath1}, leverages advanced machine learning to enable objective and quantitative analysis of whole slide images (WSIs). These methods support automated diagnosis by interpreting complex histopathological patterns. However, the extreme scale and structural heterogeneity of WSIs present major obstacles to end-to-end learning. To circumvent the need for exhaustive pixel-level annotations, weakly supervised learning has become the de facto paradigm, most notably through the multiple instance learning (MIL) framework~\cite{weak1,weak2,weak3,weak4}.

\begin{figure} 
	\centering
	\includegraphics[trim=0 0 0 0,clip,scale=0.115]{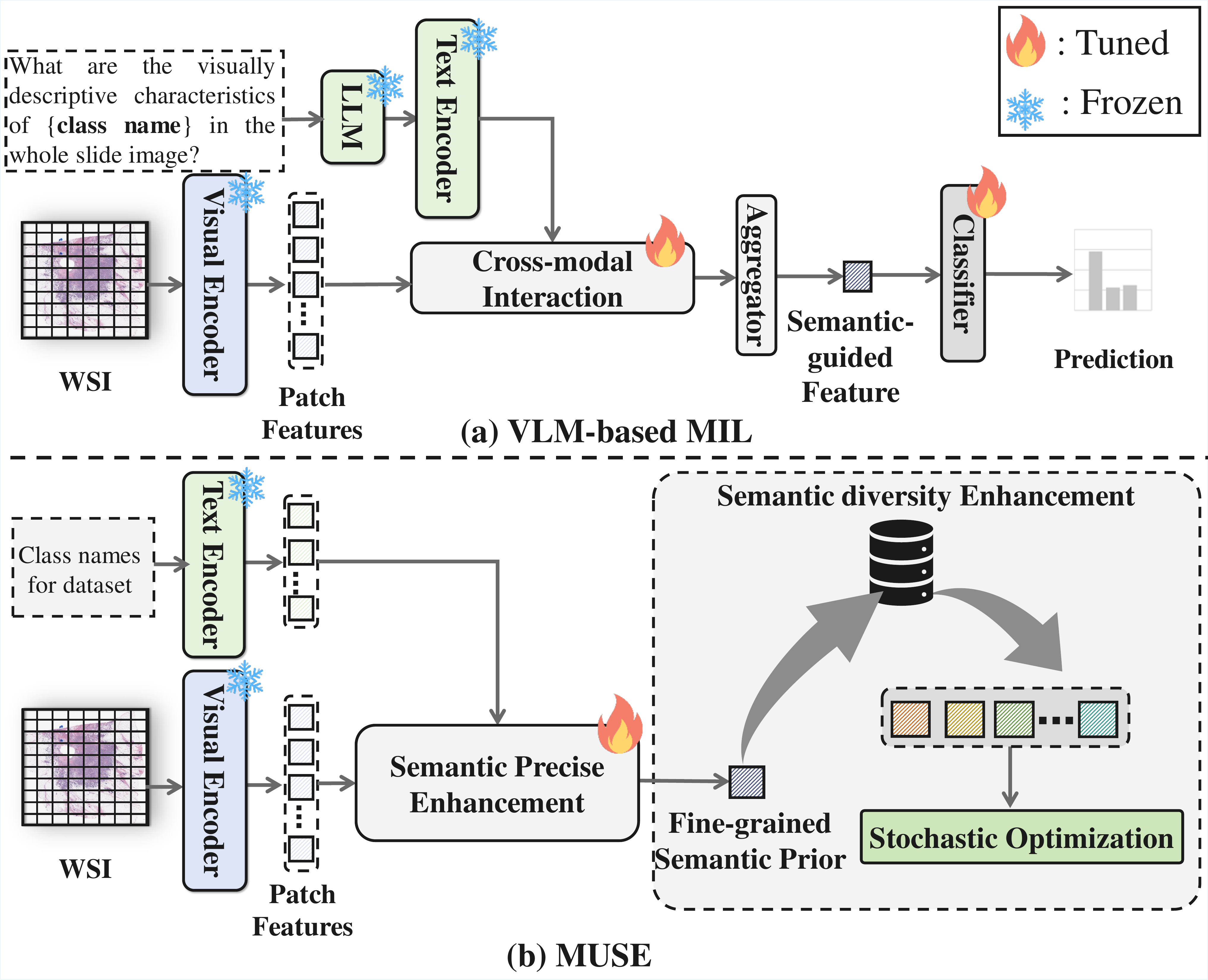} 
	\caption{Comparison of MUSE with existing VLM-based MIL methods. (a) VLM-based MIL methods incorporate pathological text and enable cross-modal interaction between text and image modalities. (b) Our method performs fine-grained modeling of semantics and enables interaction between text and image modalities, while enhancing text diversity through knowledge base retrieval and stochastic optimization.}
	\label{intro}
\end{figure}

Conventional MIL approaches for WSI analysis~\cite{mhim, sam_mil, abmil, ibmil, dtfd_mil, survival1, survival2} typically follow a three-stage pipeline: (1) patch extraction, (2) feature encoding via a pretrained backbone, and (3) bag-level aggregation under slide-level supervision. These methods have demonstrated strong performance in tasks such as cancer diagnosis~\cite{abmil, mhim}, subtyping~\cite{ibmil, dtfd_mil, sam_mil}, and survival prediction~\cite{survival1, survival2}—but only when sufficient labeled WSIs are available.

In practice, however, acquiring slide-level annotations requires expert pathologists and is further constrained by stringent data privacy regulations. Consequently, labeled WSIs are not only scarce but often limited to just a few examples per diagnostic category. This reality necessitates a few-shot learning paradigm, where models must generalize from minimal labeled slides. Under such extreme data scarcity, visual features alone are insufficient to capture the nuanced diagnostic criteria that distinguish pathological subtypes. Instead, high-level semantic knowledge such as disease descriptions, histological terminology, and clinical narratives becomes critical. Such semantics encode invariant diagnostic principles shared across patients and institutions, offering a powerful inductive bias for generalization.

Motivated by this, recent data-efficient methods~\cite{top, vila_mil, mscpt, path_enhanced, fast} leverage vision-language models (VLMs)~\cite{clip, blip} pretrained on large-scale image-text corpora to align histopathological patterns with their corresponding textual semantics. By fusing or interacting visual and semantic representations, these approaches aim to solve the few-shot WSI classification (FSWC) problem, effectively using language as a proxy for expert knowledge when labeled data is unavailable.

Recent vision-language methods for computational pathology have moved beyond generic models trained on natural images by adopting pathology-specific foundation models such as PLIP~\cite{plip}, CONCH~\cite{conch}, and MUSK~\cite{musk}. These models are pretrained on large-scale histopathology image-text pairs via contrastive learning~\cite{quilt, plip, conch, musk} and provide domain-aligned encoders for both modalities. Building on this progress, several approaches~\cite{focus, convlm, vila_mil, mgpath} leverage large language models to generate additional textual descriptions and design cross-modal interactions between text and patch features, as shown in Figure~\ref{intro}(a). However, large language models are typically used only as description generators rather than semantic optimizers, resulting in static and unrefined textual prompts.

This superficial use of semantics leads to two key limitations. First, complex pathological concepts are often collapsed into a single global query, preventing disentanglement of fine-grained diagnostic attributes such as tumor grade or immune infiltration. As a result, visual-semantic alignment remains coarse and fails to attend to diagnostically relevant regions with concept-level precision. Second, the reliance on unoptimized prompts ignores the structural diversity of clinical language, including variations in abstraction level, contextual nuance, and syntactic formulation. Under few-shot settings, this not only underutilizes the expressive capacity of the text encoder but also encourages overfitting to specific phrasings, degrading generalization across clinical contexts.

To address the aforementioned limitations, we propose a stochastic \textbf{MU}lti-view \textbf{S}emantic \textbf{E}nhancement framework, abbreviated as \textbf{MUSE}. As shown in Figure~\ref{intro}(b), the core idea of MUSE is to jointly enhance the model's generalization capability through precise semantic perception and enriched semantic diversity. Our framework consists of two core components: Sample-wise Fine-grained Semantic Enhancement (SFSE) and Stochastic Multi-view Model Optimization (SMMO). SFSE enhances semantic precision through decompositional semantic refinement and fine-grained, query-driven sample-wise cross-modal interaction. SMMO promotes semantic diversity by retrieving multi-view textual descriptions from a contextual semantic knowledge base generated by an LLM with semantics refined by SFSE, followed by a fast stochastic optimization process for final WSI classification. These components enable MUSE to achieve strong generalization in few-shot settings. Extensive experiments on multiple widely used WSI datasets show that our method achieves superior performance compared to existing FSWC baselines. The main contrutions of our paper can be summarized as follows:
\begin{itemize}    
\item We propose the MUSE framework, which improves semantic understanding in multiple instance learning through fine-grained semantic modeling and effective exploitation of semantic diversity, significantly boosting generalization in few-shot scenarios. To the best of our knowledge, this work is the first attempt to improve few-shot WSI classification performance from the perspective of semantic optimization.

\item We propose an MoE based mechanism that decompositionally refines category level semantics and adapts them to individual samples through interaction with visual features. This enables the learning of sample-wise semantic priors that capture fine-grained semantic cues, thereby enhancing semantic precision beyond conventional class level representations.

\item We build an LLM-generated knowledge base of multi-view and class-specific pathological descriptions, whose semantic diversity offers complementary signals for few-shot learning. Guided by SFSE-refined sample-wise priors and integrated stochastically, these multi-view semantics enhance generalization under limited labels.

\end{itemize}

%% file: sec/2_relatedHS.tex
\section{Related Works}
\label{sec:formatting}

\subsection{Multiple Instance Learning in CPath}
Due to the ultra-high resolution of whole-slide images (WSIs), multiple instance learning (MIL) has become the standard framework in computational pathology. MIL-based methods have demonstrated strong performance on WSI diagnosis, subtyping, and prognosis tasks \cite{abmil, ibmil, dtfd_mil, survival1, survival2}. A typical pipeline first tiles the WSI into patches, extracts features using a pretrained encoder, and then aggregates these features to predict the slide-level label.
Early aggregators relied on parameter-free pooling operations \cite{no_parm}. Subsequent works introduced learnable mechanisms to identify diagnostically relevant patches. ABMIL \cite{abmil} uses attention to assign importance scores to individual patches. CLAM \cite{clam} enhances this with clustering constraints to localize critical regions. TransMIL \cite{transmil} models global inter-patch dependencies via self-attention, while graph-based methods \cite{wikg, graph1, graph2} incorporate spatial structure to improve contextual reasoning.
However, under sparse annotations, models relying solely on visual features often underperform, highlighting the need to integrate domain knowledge for effective weakly supervised learning.

\subsection{Vision-Language Models in CPath}
General-purpose vision-language models such as CLIP \cite{clip} and BLIP \cite{blip} have demonstrated strong performance across a wide range of visual tasks \cite{dual_view, text_as_image, category, maple}. In computational pathology (CPath), domain-adapted foundation models including PLIP \cite{plip}, CONCH \cite{conch}, and MUSK \cite{musk} leverage large-scale pathological image-text data to improve diagnosis and downstream analysis.
To address sparse annotations, recent works integrate these models into the MIL framework by exploiting their few-shot and zero-shot transfer capabilities. Top \cite{top} introduces the FSWC paradigm using text-guided patch aggregation for WSI classification under data scarcity. ViLa-MIL \cite{vila_mil} proposes a dual-scale MIL framework that fuses textual descriptions with image features at multiple resolutions. FOCUS \cite{focus} enhances representation quality through a knowledge-guided adaptive visual compression mechanism.
However, these methods treat semantic prompts as static category-level descriptors, ignoring both sample-wise fine-grained semantics and the structural and perspectival diversity of clinical language, thereby limiting the expressiveness of vision-language learning.

%% file: sec/3_method_HS.tex
\section{Methodology}

\begin{figure*} 
	\centering
	\includegraphics[trim=0 0 0 0,clip,scale=0.175]{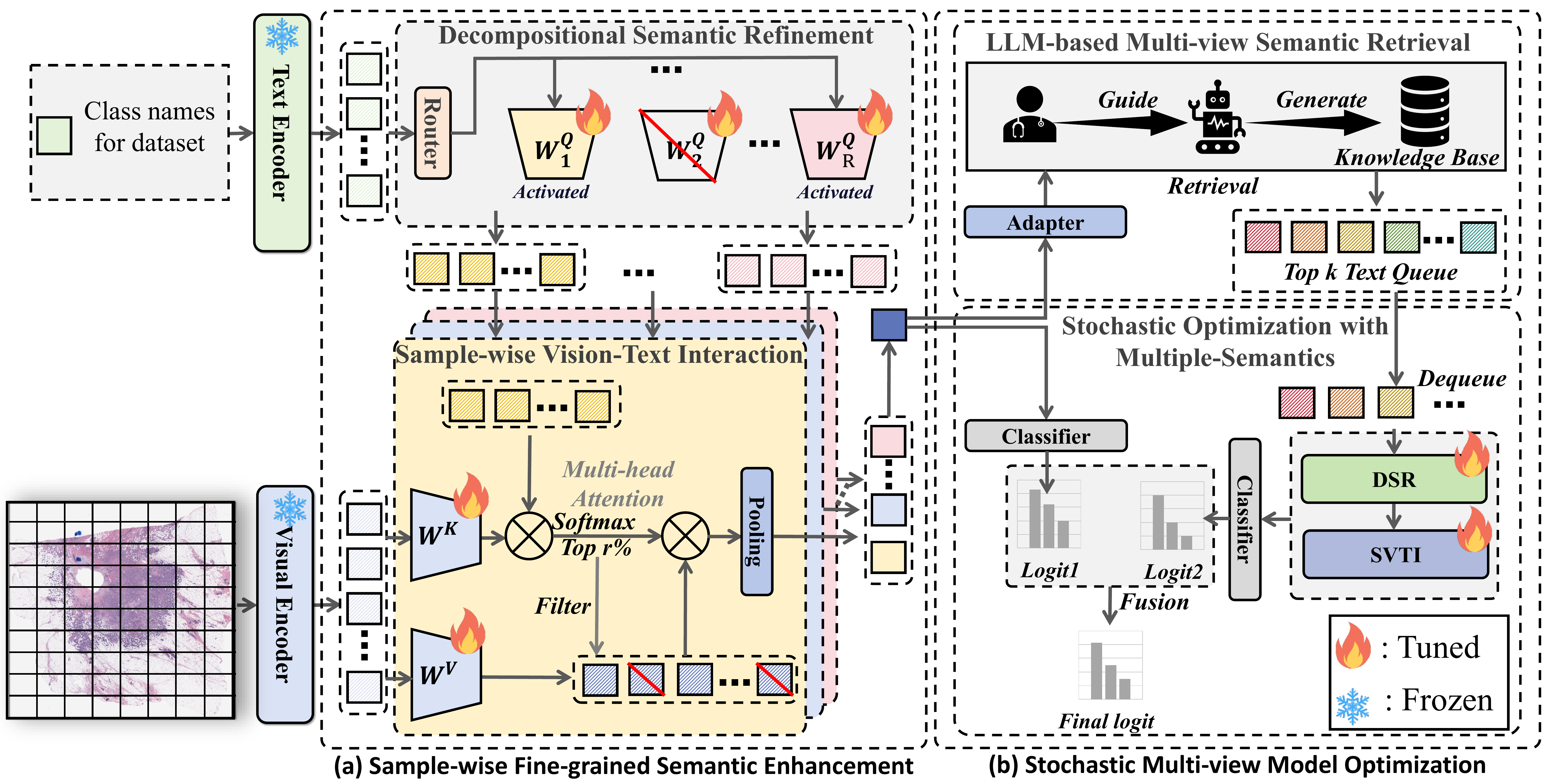} 
	\caption{Overview of the proposed MUSE framework. (DSR: Decompositional Semantic Refinement. SVTI: Sample-wise Vision-Text Interaction) (a) The input semantic information is decomposed and modeled in a fine-grained manner. We then leverage the refined semantic representations to extract sample-relevant visual-semantic information and facilitate cross-modal interaction. (b) The semantic-enhanced features are used to retrieve relevant texts from the multi-view text knowledge base, and these retrieved texts are subsequently leveraged through stochastic optimization to enrich semantic diversity.}
	\label{model}
\end{figure*}

\subsection{Overview}
We propose a stochastic multi-view semantic enhancement framework for few-shot whole slide image classification, termed MUSE, as shown in Figure~\ref{model}. The framework consists of two core components: sample-wise fine-grained semantic enhancement (SFSE) and stochastic multi-view model optimization (SMMO). Through these components, we achieve fine-grained semantic modeling and effectively harness semantic diversity, thereby enhancing the model’s understanding of the semantic modality and improving its generalization capability under few-shot settings.

\subsection{Fine-grained Semantic Enhancement}
As shown in Figure~\ref{model}(a), the sample-wise fine-grained semantic enhancement (SFSE) comprises two components: decompositional semantic refinement (DSR) and sample-wise vision–text interaction (SVTI). DSR decomposes the input textual semantics and isolates task-relevant core semantic segments, which serve as semantic cues for cross-modal interaction. SVTI employs cross-attention to dynamically attend to visual patches using these cues as queries, selectively aggregating semantically relevant features at the sample level. This attention-driven fusion enriches the representation with fine-grained, sample-wise, and context-aware semantic information.

\subsubsection{Decompositional Semantic Refinement}\label{subsec:moE}
To adapt the frozen pathology foundation model to the downstream FSWC task, we augment each category name $c \in C$ with $M$ learnable prompt vectors. For example, in CAMELYON~\cite{camelyon16,camelyon17}, categories include ``normal lymph node'' and ``metastatic lymph node''. The resulting text prompt for category $c$ is formulated as:
\begin{equation}
    T_c = [V]_1 [V]_2 \dots [V]_M [c],
\end{equation}
where $[V]_i \in \mathbb{R}^d$ ($i=1,\dots,M$) are trainable embeddings. This prompt is encoded by the text encoder $E_T(\cdot)$ of the foundation model to produce a textual feature representation $D \in \mathbb{R}^{|C| \times d}$, where each row $D_i$ corresponds to the encoded semantics of the $i$-th category.

Although prompt tuning enhances task adaptability, the resulting representation remains holistic and lacks explicit fine-grained structure. To address this, the Decompositional Semantic Refinement (DSR) module refines $D$ into task-relevant semantic cues. Specifically, inspired by the Mixture-of-Experts (MoE) paradigm, we construct $R$ expert query matrices $\{W^Q_i\}_{i=1}^R$. A lightweight router network scores each expert based on the input semantics:
\begin{equation}
    S = G(D) + \epsilon(D),
\end{equation}
where $S \in \mathbb{R}^{|C| \times R}$, $G(\cdot)$ is a linear projection, and $\epsilon(D)$ injects input-dependent Gaussian noise to encourage expert diversity and prevent collapse.

For each category $i$, we select the top-$k$ experts with the highest scores and retrieve their corresponding query matrices $\{W^Q_{ij}\}_{j=1}^k$. The semantic decomposition is then performed via:
\begin{equation}
    Q_{ij} = D_i W^Q_{ij}, \quad j = 1, \dots, k.
\end{equation}
This yields a set of refined semantic cues $\{Q_{ij}\}_{j=1}^k$ for category $i$, which serve as semantic queries for sample-wise vision--text interaction in the subsequent stage.
\subsubsection{Sample-wise Vision--Text Interaction}
For each input WSI, we first employ the visual feature extractor $E_I(\cdot)$ of the foundation model to extract patch-level features $\{b_n\}_{n=1}^{N}$ and form a bag representation $B \in \mathbb{R}^{N \times d}$.

Given the $i$-th category semantics $D_i$, we leverage the fine-grained semantic cues $\{Q_{ij}\}_{j=1}^{k}$ produced by DSR, which encode discriminative sub-concepts relevant to pathological diagnosis. The goal of SVTI is to dynamically identify and fuse the subset of cues most aligned with the visual content of the current sample through vision--text interaction, thereby constructing a sample-level semantic prior.

Specifically, we compute multi-head cross-attention between each cue $Q_{ij}$ and the visual bag $B$. For attention head $h$, the alignment scores are:
\begin{equation}
    A^{h}_{ij} = \frac{(Q_{ij} W^{Q,h}) (B W^{K,h})^\top}{\sqrt{d_{\text{head}}}},
\end{equation}
where $W^{Q,h}, W^{K,h} \in \mathbb{R}^{d \times d_{\text{head}}}$ are learnable projection matrices.

To focus on regions highly correlated with the semantic cues, we retain only the top-$r\%$ patches with the highest attention scores for each head. Let $\mathcal{I}_{ij}^h$ denote the corresponding index set. The filtered value projection is then:
\begin{equation}
    V^{h,\text{filtered}}_{ij} = B_{\mathcal{I}_{ij}^h} W^{V,h},
\end{equation}
with $W^{V,h} \in \mathbb{R}^{d \times d_{\text{head}}}$ as the value projection matrix.

We aggregate across heads to obtain a vision--text fused representation for cue $j$:
\begin{gather}
    \text{head}_h = \text{Softmax}(A^{h}_{ij}) \, V^{h,\text{filtered}}_{ij}, \\
    f_{ij} = \text{Concat}(\text{head}_1, \dots, \text{head}_H) W^{O},
\end{gather}
where $W^{O} \in \mathbb{R}^{H d_{\text{head}} \times d}$ is the output projection matrix.

Finally, we combine the $k$ fused representations using the expert scores $\{S_{ij}\}_{j=1}^k$ from the DSR router, and obtain the final representation $f$ through average pooling:
\begin{gather}
    \{S_{ij}^{\text{norm}}\}_{j=1}^{k} = \text{Softmax}(\{S_{ij}\}_{j=1}^{k}), \\
   f_i = \sum_{j=1}^{k} S_{ij}^{\text{norm}} f_{ij},  \quad
   f = \frac{1}{\left| C \right|}\sum_{i=1}^{\left| C \right|}f_i.
\end{gather}
The resulting vector $f$ serves as a fine-grained semantic prior for the input WSI with respect to the corresponding text feature $D$. This prior reflects the semantic sub-concepts most supported by the current sample visually, and is specifically designed to serve as a query for retrieving complementary textual knowledge from an external semantic knowledge base in the subsequent stage.

\subsection{Stochastic Multi-view Model Optimization}
As illustrated in Figure~\ref{model}, the sample-specific semantic prior $f$ generated by SVTI is used to query a pathology-oriented knowledge base, which is constructed offline via LLM-guided generation using chain-of-thought (CoT)~\cite{cot} and in-context learning (ICL)~\cite{icl} to produce diverse, multi-view textual descriptions. During training, the stochastic multi-view model optimization retrieves a set of semantically complementary texts using $f$ and randomly samples one of them at each iteration to update the model, thereby improving generalization through exposure to diverse semantic views.

\subsubsection{LLM-based Semantic Knowledge Base Generation} \label{subsec:pipeline}
\begin{figure} 
	\centering
	\includegraphics[trim=0 0 0 0,clip,scale=0.108]{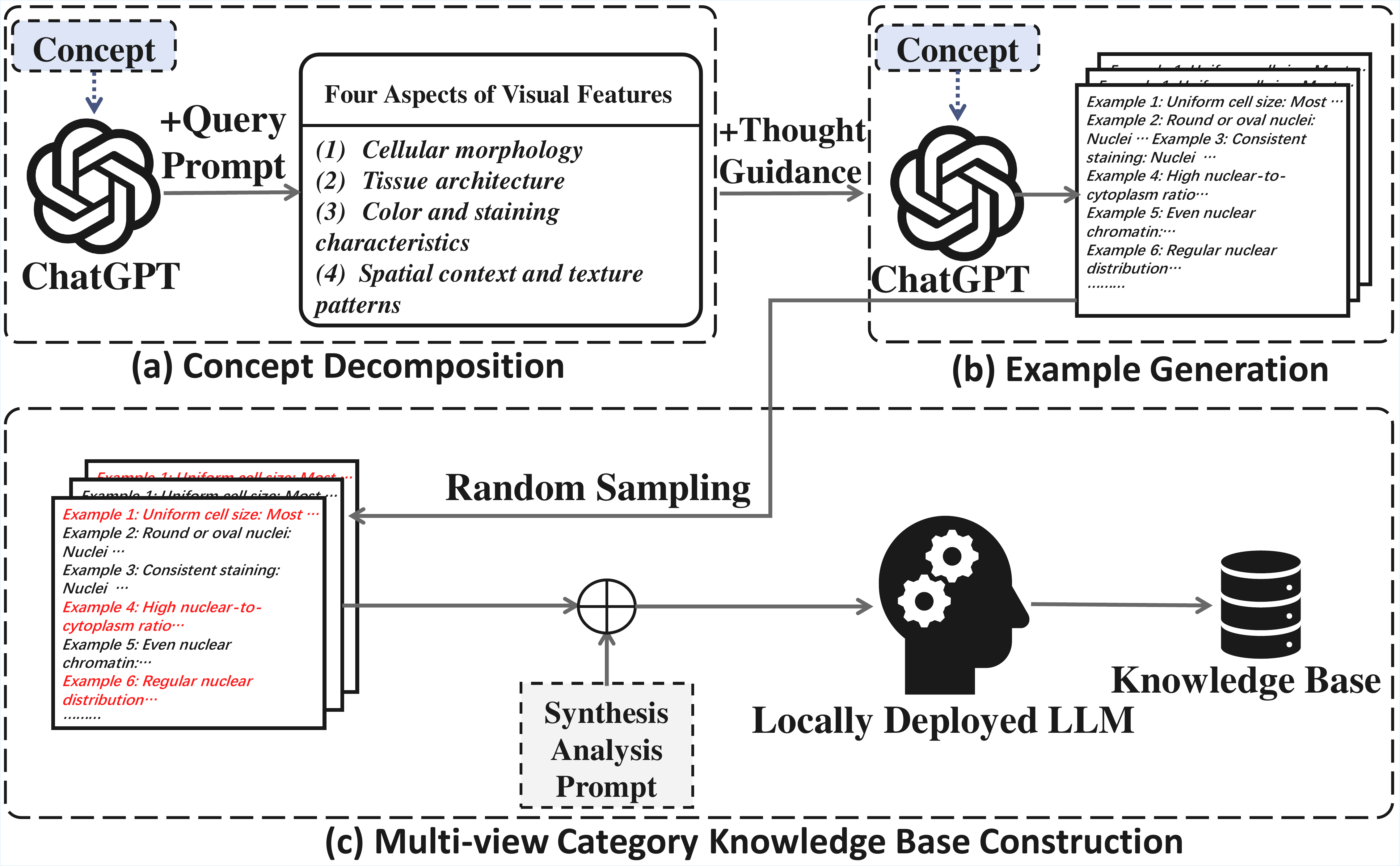} 
	\caption{Our pipeline of the category related text knowledge base generation. (a) We employ ChatGPT to analyze and decompose the concepts associated with class names. (b) We leverage ChatGPT to construct concrete examples for each aspect. (c) We randomly sample some examples and combine them with prompts to guide the locally deployed LLM in generating the category knowledge base.}
	\label{LLM_pipeline}
\end{figure}

As shown in Figure~\ref{LLM_pipeline}, we construct a multi-view textual knowledge base for each pathological category $c$ by leveraging category-level diagnostic concepts to guide LLM generation. The process involves three steps: concept decomposition, exemplar generation, and knowledge base assembly.

First, we prompt GPT-4 with the category name to decompose its visual diagnostic criteria into four clinically meaningful aspects: cellular morphology, tissue architecture, color-staining characteristics, and spatial-texture patterns (Figure~\ref{LLM_pipeline}(a)).

Next, for each aspect and category, GPT-4 generates 10 concrete exemplar descriptions (Figure~\ref{LLM_pipeline}(b)). To promote semantic diversity, we form synthesis prompts by randomly selecting one exemplar from each aspect and concatenating them. Due to API cost and time constraints, we use a lightweight open-source LLM deployed locally to generate 300 multi-view descriptions per category based on these prompts (Figure~\ref{LLM_pipeline}(c)).

Finally, all generated texts are encoded by the text encoder $E_T(\cdot)$ into feature vectors, forming the category-specific knowledge base $\mathcal{B}_c$. Additional details are provided in the Supplementary Material.

\subsubsection{LLM-based Multi-view Semantic Retrieval}
Given the sample-wise semantic prior $f$ generated by SVTI and the ground-truth category label $c$, we retrieve semantically relevant text features from the pre-constructed knowledge base $\mathcal{B}_c$. The retrieval is performed by computing cosine similarity between an adapted representation of $f$ and all entries in $\mathcal{B}_c$:
\begin{equation}
T_m = \operatorname{Top\text{-}m}_{\,t \in \mathcal{B}_c} \big( \text{Sim}(A(f), t) \big),
\end{equation}
where $A(\cdot)$ is a lightweight adapter that aligns the feature space of $f$ with that of the text encoder, and $\text{Sim}(\cdot,\cdot)$ denotes cosine similarity.

To facilitate subsequent stochastic utilization, the retrieved set $T_m$ is randomly shuffled and stored in a queue structure, enabling sequential popping of text features in a randomized order during optimization.

\subsubsection{Stochastic Optimization with Multiple-Semantics}\label{sec:soms}
Building upon the retrieved text features $T_m$ , we stochastically incorporate diverse textual views to enrich semantic representation. The set $T_m$ is stored in a shuffled queue to enable random access during training.

At each iteration, a single text feature $t$ is dequeued from $T_m$. We apply Decompositional Semantic Refinement (DSR) to decompose $t$ into fine-grained semantic queries $\{Q^{t}_{j}\}_{j=1}^{k}$, which are then processed by the Sample-wise Vision-Text Interaction (SVTI) module together with the input bag $B$ to produce an auxiliary sample-wise prior $f^{t}_{\text{aux}}$. This prior introduces semantic information from additional textual views, thereby compensating for the semantic representation of the primary prior $f$.

Both $f$ and $f^{t}_{\text{aux}}$ are projected into logits via a shared MLP:
\begin{gather}
z = \text{MLP}(f), \quad
z^{t}_{\text{aux}} = \text{MLP}(f^{t}_{\text{aux}}),
\end{gather}
and fused by summation:
\begin{equation}
z^{t}_{\text{final}} = \frac{z + z^{t}_{\text{aux}}}{2}.
\end{equation}
The model is trained with cross-entropy loss:
\begin{equation}
\mathcal{L}^{t} = \text{CE}(z^{t}_{\text{final}}, GT),
\end{equation}
where $GT$ denotes the sample-wise (slide-level) ground-truth label.

By stochastically exposing the model to multiple textual views from $T_m$, this procedure realizes multi-view semantic compensation, thereby improving generalization in few-shot scenarios. The full training algorithm is provided in the Supplementary Material.

%% file: sec/4_experiment.tex
\section{Experiments}

\begin{table*}
\centering
\caption{Few-shot weakly supervised learning results (presented in \%) on CAMELYON, TCGA-NSCLC, and TCGA-BRCA under 4-shot, 8-shot, and 16-shot settings are presented. The best performance is highlighted in \textbf{bold}, and the second-best is \underline{underlined}.}
\scriptsize
\renewcommand{\arraystretch}{1}
\setlength{\tabcolsep}{3.5pt}
\resizebox{\linewidth}{!}{ 
\begin{tabular}{c|c|ccc|ccc|ccc}
\toprule
\multirow{3}{*}{\textbf{Dataset}} &  \multirow{3}{*}{\textbf{Model}} & \multicolumn{3}{c|}{\textbf{4-shot \rule{0pt}{1.8ex}}} & \multicolumn{3}{c|}{\textbf{8-shot \rule{0pt}{0ex}}} & \multicolumn{3}{c}{\textbf{16-shot \rule{0pt}{0ex}}} \\ \cmidrule(l){3-11}
& & ACC & AUC & F1 Score & ACC & AUC & F1 Score & ACC & AUC & F1 Score \\
\midrule
\multirow{12}{*}{\rotatebox[origin=c]{90}{\shortstack{\normalsize\textbf{CAMELYON}}}}
& Mean Pooling & 59.07 \tiny±3.03 & 51.19 \tiny±4.81 & 47.17 \tiny±5.19 & 64.05 \tiny±4.14 & 61.53 \tiny±5.07 & 55.14 \tiny±2.90 & 65.83 \tiny±1.96 & 62.50 \tiny±4.56 & 55.04 \tiny±7.53 \\

& Max Pooling & 62.93 \tiny±4.20 & 62.50 \tiny±8.88 & 55.58 \tiny±7.61 & 70.37 \tiny±5.00 & 74.63 \tiny±4.38 & 67.64 \tiny±4.44 & 74.46 \tiny±6.46 & 78.38 \tiny±8.12 & 72.15 \tiny±6.45 \\

& ABMIL \cite{abmil} & 62.04 \tiny±2.77 & 58.98 \tiny±7.37 & 49.61 \tiny±8.85 & 71.52 \tiny±10.41 & 75.05 \tiny±11.03 & 69.69 \tiny±10.15 & 82.37 \tiny±7.10 & 85.48 \tiny±6.22 & 81.04 \tiny±7.36 \\

& DSMIL \cite{dsmil} & 59.44 \tiny±9.00 & 54.45 \tiny±10.00 & 53.01 \tiny±9.16 & 73.19 \tiny±6.93 & 74.63 \tiny±9.69 & 70.20 \tiny±8.79 & 84.72 \tiny±5.11 & 87.80 \tiny±4.69 & 83.60 \tiny±5.35 \\

& CLAM-SB \cite{clam} & 57.59 \tiny±6.18 & 55.42 \tiny±9.81 & 53.67 \tiny±6.09 & 71.67 \tiny±7.50 & 73.56 \tiny±10.40 & 67.16 \tiny±10.43 & 81.30 \tiny±9.97 & 84.65 \tiny±10.30 & 79.91 \tiny±10.34 \\

& CLAM-MB \cite{clam} & 60.22 \tiny±4.56 & 54.90 \tiny±7.69 & 50.38 \tiny±7.11 & 72.41 \tiny±10.11 & 73.88 \tiny±12.15 & 68.48 \tiny±12.69 & 80.03 \tiny±8.03 & 84.01 \tiny±9.48 & 78.13 \tiny±9.61 \\

& TransMIL \cite{transmil} & 58.95 \tiny±4.08 & 52.07 \tiny±3.26 & 47.01 \tiny±3.34 & 66.91 \tiny±2.71 & 62.72 \tiny±7.56 & 55.27 \tiny±5.33 & 68.81 \tiny±6.13 & 68.24 \tiny±10.04 & 61.23 \tiny±11.19 \\

& WiKG \cite{wikg} & 62.15 \tiny±3.48 & 56.60 \tiny±6.96 & 51.13 \tiny±5.53 & 65.98 \tiny±3.63 & 65.10 \tiny±6.05 & 57.84 \tiny±6.17 & 71.00 \tiny±6.90 & 70.73 \tiny±10.08 & 66.58 \tiny±8.68 \\

& Top \cite{top} & 63.04 \tiny±3.99 & 56.39 \tiny±7.22 & 49.28 \tiny±7.43 & 66.39 \tiny±4.36 & 64.98 \tiny±7.13 & 54.85 \tiny±10.01 & 77.54 \tiny±9.07 & 76.76 \tiny±12.51 & 72.11 \tiny±13.26 \\

& ViLa-MIL \cite{vila_mil} & 65.20 \tiny±5.30 & 62.96 \tiny±8.16 & 53.91 \tiny±11.15 & 76.87 \tiny±7.17 & 79.15 \tiny±8.19 & 72.79 \tiny±9.30 & 84.60 \tiny±4.27 & 88.00 \tiny±4.23 & 83.24 \tiny±4.55 \\

& FOCUS \cite{focus} & \underline{68.13 \tiny±8.16} & \underline{68.84 \tiny±10.32} & \underline{63.40 \tiny±7.85} & \underline{80.33 \tiny±8.64} & \underline{83.95 \tiny±9.03} & \underline{78.19 \tiny±10.51} & \underline{88.62 \tiny±1.96} & \underline{91.51 \tiny±2.38} & \underline{87.49 \tiny±2.19} \\

& \cellcolor{gray!15}MUSE  
& \cellcolor{gray!15}\textbf{74.86 \tiny±9.22} & \cellcolor{gray!15}\textbf{76.65 \tiny±14.92} & \cellcolor{gray!15}\textbf{68.66 \tiny±15.50}
& \cellcolor{gray!15}\textbf{84.01 \tiny±4.46} & \cellcolor{gray!15}\textbf{88.32 \tiny±4.29} & \cellcolor{gray!15}\textbf{82.42 \tiny±4.62}
& \cellcolor{gray!15}\textbf{89.70 \tiny±1.57} & \cellcolor{gray!15}\textbf{92.52 \tiny±1.57} & \cellcolor{gray!15}\textbf{88.59 \tiny±1.91} \\
\hline

\multirow{12}{*}{\rotatebox[origin=c]{90}{\shortstack{\normalsize\textbf{TCGA-NSCLC}}}}

& Mean Pooling & 71.32 \tiny±7.78 & 77.87 \tiny±10.46 & 70.72 \tiny±8.55 & 78.13 \tiny±4.14 & 85.89 \tiny±4.61 & 78.06 \tiny±4.16 & 83.60 \tiny±3.97 & 91.44 \tiny±3.09 & 83.51 \tiny±4.05 \\

& Max Pooling & 68.22 \tiny±8.41 & 75.22 \tiny±10.51 & 67.54 \tiny±9.03 & 76.64 \tiny±6.92 & 83.27 \tiny±7.42 & 76.58 \tiny±6.93 & 85.28 \tiny±3.07 & 92.93 \tiny±2.74 & 85.25 \tiny±3.08 \\

& ABMIL \cite{abmil} & 78.38 \tiny±8.60 & 84.93 \tiny±9.47 & 78.25 \tiny±8.68 & 85.88 \tiny±4.40 & 93.33 \tiny±3.85 & 85.84 \tiny±4.42 & \underline{89.24 \tiny±2.76} & \underline{96.08 \tiny±1.31} & \underline{89.22 \tiny±2.76} \\

& DSMIL \cite{dsmil} & 77.31 \tiny±9.08 & 84.77 \tiny±10.67 & 77.19 \tiny±9.16 & 85.44 \tiny±4.90 & 92.33 \tiny±3.75 & 85.42 \tiny±4.90 & 89.14 \tiny±2.27 & 95.66 \tiny±1.58 & 89.12 \tiny±2.27 \\

& CLAM-SB \cite{clam} & 76.45 \tiny±8.63 & 84.15 \tiny±10.68 & 76.05 \tiny±9.36 & 85.28 \tiny±3.57 & 92.17 \tiny±3.78 & 85.25 \tiny±3.58 & 88.73 \tiny±4.12 & 95.38 \tiny±2.67 & 88.68 \tiny±4.20 \\

& CLAM-MB \cite{clam} & 78.00 \tiny±6.81 & 86.69 \tiny±7.49 & 77.73 \tiny±7.03 & 84.36 \tiny±3.47 & 91.79 \tiny±3.01 & 84.31 \tiny±3.50 & 87.87 \tiny±2.80 & 95.27 \tiny±1.75 & 87.84 \tiny±2.83 \\

& TransMIL \cite{transmil} & 71.93 \tiny±8.35 & 78.54 \tiny±9.95 & 71.19 \tiny±9.54 & 80.34 \tiny±4.14 & 87.78 \tiny±4.18 & 80.26 \tiny±4.21 & 86.51 \tiny±2.85 & 93.98 \tiny±2.26 & 86.49 \tiny±2.86 \\

& WiKG \cite{wikg} & 71.86 \tiny±6.37 & 80.80 \tiny±7.08 & 71.43 \tiny±6.82 & 78.48 \tiny±4.09 & 86.43 \tiny±4.83 & 78.36 \tiny±4.14 & 84.43 \tiny±3.80 & 92.24 \tiny±3.08 & 84.40 \tiny±3.80 \\

& Top \cite{top} & 69.74 \tiny±7.97 & 74.41 \tiny±9.98 & 69.37 \tiny±8.33 & 81.44 \tiny±3.27 & 83.84 \tiny±9.17 & 70.29 \tiny±9.34 & 84.30 \tiny±5.14 & 91.68 \tiny±4.14 & 84.26 \tiny±5.15 \\

& ViLa-MIL \cite{vila_mil}  & 74.96 \tiny±8.36 & 82.80 \tiny±10.16 & 74.55 \tiny±8.90 & 83.57 \tiny±5.04 & 91.00 \tiny±4.97 & 83.49 \tiny±5.05 & 88.41 \tiny±3.50 & 95.12 \tiny±2.67 & 88.37 \tiny±3.56 \\

& FOCUS \cite{focus} & \underline{79.14 \tiny±6.96} & \underline{87.04 \tiny±8.06} & \underline{78.96 \tiny±7.32} & \underline{86.04 \tiny±3.37} & \underline{93.69 \tiny±2.30} & \underline{85.96 \tiny±3.48} & 88.38 \tiny±3.02 & 95.63 \tiny±1.67 & 88.35 \tiny±3.05 \\

& \cellcolor{gray!15}MUSE  
& \cellcolor{gray!15}\textbf{79.90 \tiny±7.23} & \cellcolor{gray!15}\textbf{87.57 \tiny±7.94} & \cellcolor{gray!15}\textbf{79.85 \tiny±7.23}
& \cellcolor{gray!15}\textbf{87.27 \tiny±3.22} & \cellcolor{gray!15}\textbf{94.27 \tiny±2.59} & \cellcolor{gray!15}\textbf{87.20 \tiny±3.29}
& \cellcolor{gray!15}\textbf{89.74 \tiny±2.87} & \cellcolor{gray!15}\textbf{96.82 \tiny±1.38} & \cellcolor{gray!15}\textbf{89.70 \tiny±2.90} \\
\hline

\multirow{12}{*}{\rotatebox[origin=c]{90}{\shortstack{\normalsize\textbf{TCGA-BRCA}}}}

& Mean Pooling & 76.65 \tiny±6.78 & 75.94 \tiny±19.57 & 66.90 \tiny±10.54 & 79.48 \tiny±5.93 & 82.19 \tiny±7.98 & 71.03 \tiny±7.89 & 84.63 \tiny±3.65 & 89.60 \tiny±2.45 & 77.38 \tiny±3.94 \\

& Max Pooling & 77.17 \tiny±5.12 & 69.55 \tiny±13.59 & 60.51 \tiny±8.24 & 76.55 \tiny±6.25 & 76.07 \tiny±6.41 & 64.63 \tiny±7.32 & 80.54 \tiny±5.87 & 84.89 \tiny±6.01 & 72.82 \tiny±6.50 \\

& ABMIL \cite{abmil} & 79.42 \tiny±6.98 & 83.43 \tiny±11.34 & 71.75 \tiny±8.94 & 79.22 \tiny±4.45 & 85.90 \tiny±6.69 & 72.39 \tiny±5.24 & 86.39 \tiny±2.86 & 91.87 \tiny±2.23 & 80.69 \tiny±3.37 \\

& DSMIL \cite{dsmil} & 78.71 \tiny±11.11 & 83.94 \tiny±9.66 & 68.62 \tiny±12.23 & 82.63 \tiny±4.61 & 82.58 \tiny±10.35 & 72.51 \tiny±9.36 & \underline{88.03 \tiny±2.67} & 91.05 \tiny±3.50 & \underline{81.38 \tiny±4.35} \\

& CLAM-SB \cite{clam} & 79.80 \tiny±12.21 & 84.96 \tiny±12.08 & 73.40 \tiny±12.41 & 81.92 \tiny±5.63 & \underline{87.24 \tiny±5.32} & \underline{74.69 \tiny±6.44} & 85.85 \tiny±3.64 & 92.03 \tiny±1.78 & 80.05 \tiny±4.04 \\

& CLAM-MB \cite{clam} & 80.70 \tiny±6.33 & \underline{86.07 \tiny±6.84} & \underline{73.43 \tiny±6.45} & 80.70 \tiny±6.33 & 86.07 \tiny±6.84 & 73.43 \tiny±6.45 & 86.94 \tiny±2.16 & \textbf{92.59 \tiny±1.77} & 81.32 \tiny±2.61 \\

& TransMIL \cite{transmil} & 79.93 \tiny±4.73 & 77.06 \tiny±10.51 & 63.71 \tiny±11.39 & 80.90 \tiny±6.88 & 81.07 \tiny±12.36 & 69.15 \tiny±11.25 & 83.89 \tiny±2.49 & 89.61 \tiny±2.92 & 75.88 \tiny±4.50 \\

& WiKG \cite{wikg} & 79.83 \tiny±3.41 & 83.79 \tiny±6.00 & 67.51 \tiny±8.66 & 81.44 \tiny±4.28 & 81.28 \tiny±8.40 & 68.63 \tiny±10.19 & 85.91 \tiny±3.21 & 90.48 \tiny±2.62 & 77.10 \tiny±9.43 \\

& Top \cite{top} & 80.96 \tiny±3.71 & 78.45 \tiny±12.48 & 65.73 \tiny±10.86 & 81.44 \tiny±3.27 & 83.38 \tiny±9.17 & 70.29 \tiny±9.34 & 85.65 \tiny±3.66 & 89.50 \tiny±3.62 & 75.54 \tiny±11.25 \\

& ViLa-MIL \cite{vila_mil} & \underline{81.80 \tiny±4.82} & 82.62 \tiny±6.54 & 68.90 \tiny±8.35 & \underline{83.92 \tiny±4.23} & 84.54 \tiny±10.03 & 72.25 \tiny±11.97 & 87.78 \tiny±2.77 & \underline{92.36 \tiny±2.32} & 81.31 \tiny±3.82 \\

& FOCUS \cite{focus} & 80.16 \tiny±8.08 & 82.81 \tiny±9.74 & 68.76 \tiny±12.63 & 83.02 \tiny±3.03 & 84.28 \tiny±8.07 & 68.94 \tiny±13.01 & 86.78 \tiny±2.52 & 92.14 \tiny±2.07 & 80.51 \tiny±2.52  \\

& \cellcolor{gray!15}MUSE  
& \cellcolor{gray!15}\textbf{84.14 \tiny±2.55} & \cellcolor{gray!15}\textbf{86.66 \tiny±8.66} & \cellcolor{gray!15}\textbf{73.81 \tiny±10.67}
& \cellcolor{gray!15}\textbf{84.37 \tiny±3.97} & \cellcolor{gray!15}\textbf{88.39 \tiny±5.17} & \cellcolor{gray!15}\textbf{76.29 \tiny±6.58}
& \cellcolor{gray!15}\textbf{88.23 \tiny±2.03} & \cellcolor{gray!15} 91.82 \tiny±2.60 & \cellcolor{gray!15}\textbf{81.44 \tiny±3.40} \\

\bottomrule
\end{tabular}
}
\label{tab:main_results}
\end{table*}

\begin{table}
\centering
\caption{Ablation experiments (presented in \%) of key modules of MUSE on on CAMELYON under few-shot learning settings.(BM: Base MIL. TI: Traditional Interaction.)}
\label{tab:dataset_summary}
\scriptsize
\resizebox{\linewidth}{!}{
\begin{tabular}{c|cccc|ccc}
\toprule
\textbf{} & \textbf{BM} & \textbf{TI} & \textbf{SFSE} & \textbf{SMMO} & \textbf{ACC} & \textbf{AUC} & \textbf{F1 Score} \\
\midrule
\multirow{5}{*}{\rotatebox{90}{\shortstack{\textbf{16-shot}}}}
& $\checkmark$ & $\times$ & $\times$ & $\times$ 
                            & 82.37 & 85.48 & 81.04\\

                            & $\checkmark$ & $\checkmark$ & $\times$ & $\times$ 
                            & 86.61 & 90.13 & 85.41\\

                            & $\checkmark$ & $\checkmark$ & $\checkmark$ & $\times$
                            & 86.72 & 90.20 & 85.33\\

                            & $\checkmark$ & $\checkmark$ & $\times$ & $\checkmark$
                            & 87.99 & 90.95 & 86.70\\

                            & \cellcolor{gray!15}$\checkmark$ & \cellcolor{gray!15}$\checkmark$ & \cellcolor{gray!15}$\checkmark$ & \cellcolor{gray!15}$\checkmark$ 
                            & \cellcolor{gray!15}\textbf{89.70} & \cellcolor{gray!15}\textbf{92.52} & \cellcolor{gray!15}\textbf{88.59}\\
\midrule
\multirow{5}{*}{\rotatebox{90}{\shortstack{\textbf{8-shot}}}}
& $\checkmark$ & $\times$ & $\times$ & $\times$ 
                            & 71.52 & 75.05 & 69.69\\

                            & $\checkmark$ & $\checkmark$ & $\times$ & $\times$ 
                            & 79.10 & 82.48 & 77.04\\

                            & $\checkmark$ & $\checkmark$ & $\checkmark$ & $\times$
                            & 80.85 & 84.05 & 78.55\\

                            & $\checkmark$ & $\checkmark$ & $\times$ & $\checkmark$ 
                            & 83.68 & 88.30 & 82.08\\

                            & \cellcolor{gray!15}$\checkmark$ & \cellcolor{gray!15}$\checkmark$ & \cellcolor{gray!15}$\checkmark$ & \cellcolor{gray!15}$\checkmark$ 
                            & \cellcolor{gray!15}\textbf{84.01} & \cellcolor{gray!15}\textbf{88.32} & \cellcolor{gray!15}\textbf{82.42}\\
\midrule
\multirow{5}{*}{\rotatebox{90}{\shortstack{\textbf{4-shot}}}}
& $\checkmark$ & $\times$ & $\times$ & $\times$ 
                            & 62.04 & 58.98 & 49.61\\

                            & $\checkmark$ & $\checkmark$ & $\times$ & $\times$ 
                            & 65.13 & 65.84 & 60.40\\

                            & $\checkmark$ & $\checkmark$ & $\checkmark$ & $\times$
                            & 65.16 & 67.30 & 60.43\\

                            & $\checkmark$ & $\checkmark$ & $\times$ & $\checkmark$ 
                            & 71.18 & 72.75 & 63.09\\

                            & \cellcolor{gray!15}$\checkmark$ & \cellcolor{gray!15}$\checkmark$ & \cellcolor{gray!15}$\checkmark$ & \cellcolor{gray!15}$\checkmark$ 
                            & \cellcolor{gray!15}\textbf{74.86} & \cellcolor{gray!15}\textbf{76.65} & \cellcolor{gray!15}\textbf{68.66}\\
\bottomrule
\end{tabular}
}
\label{tab:component_ablation}
\end{table}

\subsection{Experimental Settings}

\subsubsection{Datasets}

To validate the effectiveness of MUSE, we conduct experiments on three datasets: CAMELYON~\cite{camelyon16,camelyon17}, TCGA-NSCLC~\cite{tcga}, and TCGA-BRCA~\cite{tcga}. CAMELYON, designed for breast cancer metastasis detection, consists of 577 normal slides and 341 slides containing metastases. TCGA-NSCLC contains slides from two lung cancer subtypes: lung adenocarcinoma (LUAD, 541 slides) and lung squamous cell carcinoma (LUSC, 512 slides). TCGA-BRCA contains two breast cancer subtypes: invasive ductal carcinoma (IDC, 826 slides) and invasive lobular carcinoma (ILC, 211 slides). Each dataset is split into training, validation, and test sets with a 6:2:2 ratio. In the few-shot setting, we randomly sample $k$ slides per class from the training set, with $k = 4, 8, 16$.

\subsubsection{Implementation Details}

The original whole slide image (WSI) at 40$\times$ magnification is processed by the CLAM toolset~\cite{clam} into 512$\times$512-pixel patches. We utilize the text encoder and visual encoder of CONCH~\cite{conch} to extract textual features and patch features, respectively. To construct the multi-perspective textual knowledge base, we first employ GPT-4~\cite{gpt4} for concept decomposition and example generation, and then build the knowledge base using a locally deployed Qwen-7B model~\cite{qwen}. All experiments were conducted using PyTorch~\cite{pytorch} on a single NVIDIA RTX 3090 GPU. Additional implementation details are provided in the Supplementary Material.

\subsubsection{Evaluation Metrics}
We use three evaluation metrics to assess model performance: accuracy (ACC), area under the receiver operating characteristic curve (AUC), and F1 score. To mitigate the impact of dataset split variability in few-shot settings, we repeat each experiment ten times and report the results as mean $\pm$ standard deviation.

\subsection{Main Results}

We evaluate MUSE under few-shot settings with 4, 8, and 16 labeled slides per class on three histopathology datasets: CAMELYON, TCGA-NSCLC, and TCGA-BRCA. We compare against a comprehensive suite of state-of-the-art methods, including traditional MIL approaches (ABMIL~\cite{abmil}, DSMIL~\cite{dsmil}, CLAM-SB/MB~\cite{clam}, TransMIL~\cite{transmil}, WiKG~\cite{wikg}), vision-language models (VLMs) (Top~\cite{top}, ViLa-MIL~\cite{vila_mil}, FOCUS~\cite{focus}), and simple aggregation baselines (mean/max pooling). All methods use CONCH~\cite{conch} features for fair comparison. As reported in Table~\ref{tab:main_results}, MUSE achieves consistently strong performance and establishes new state-of-the-art results across most settings.
Notably, MUSE exhibits the largest gains when labeled data is most limited. In the 4-shot setting, it improves ACC over the best baseline by 6.73\%, 0.76\%, and 2.34\% on CAMELYON, TCGA-NSCLC, and TCGA-BRCA, respectively. In the 8-shot setting, MUSE outperforms the best baseline in F1-score by 4.23\%, 1.24\%, and 1.60\% on the same datasets. Across all benchmarks, the performance gap between MUSE and existing methods widens as the number of shots decreases. This confirms that our framework effectively leverages semantic priors to mitigate the scarcity of visual supervision, which is a key challenge in medical few-shot learning.
We further observe that VLM-based methods generally surpass conventional MIL approaches under low-shot conditions, highlighting the value of semantic information in whole-slide image classification. Crucially, MUSE consistently outperforms all VLM baselines across all metrics in both 4-shot and 8-shot settings, validating the effectiveness of our proposed semantic enhancement.

\subsection{Ablation Studies}

\subsubsection{Component Analysis}
To evaluate the contribution of each module in MUSE to the overall performance, we conduct an ablation study on CAMELYON using four variants under 4, 8, and 16-shot settings. The results are reported in Table~\ref{tab:component_ablation}. 
BM (Base MIL) denotes classification based solely on aggregated patch features. TI (Traditional Interaction) refers to feature aggregation after cross-attention between textual prompts and patch features. SFSE and SMMO correspond to the modules shown in Figure~\ref{model} (a) and (b), which perform semantic fine-grained modeling and semantic diversity enhancement, respectively. 
The results show that performance improves incrementally as each component is added. Notably, the SMMO module yields substantial gains under the 4 and 8-shot settings, highlighting its critical role in low-data regimes.

\subsubsection{Retrieval Strategy Ablation}
To investigate the impact of retrieval strategies on performance, we evaluate three approaches: Random, L2-norm, and Cosine. 
Random selects text entries uniformly at random from the knowledge base. 
L2-norm and Cosine retrieve the nearest text entries based on Euclidean distance and cosine similarity in the feature space, respectively. 
Results in Table~\ref{tab:retrieval_ablation} show that similarity-based retrieval consistently outperforms random selection. 
Moreover, cosine similarity yields the best performance under both 4 and 16-shot settings, demonstrating its effectiveness for semantic alignment in few-shot WSI classification.

\begin{table}
\centering
\caption{Performance comparison (presented in \%) of different retrieval methods in SMMO on CAMELYON under few-shot learning settings.}
\scriptsize
\resizebox{\linewidth}{!}{
\begin{tabular}{c|c|ccc}
\toprule
\textbf{} & \textbf{Retrieval Method} & \textbf{ACC} & \textbf{AUC} & \textbf{F1 Score} \\
\midrule
\multirow{3}{*}{\rotatebox{90}{\shortstack{\textbf{16-shot}}}} 
& Random & 88.17 & 90.86 & 87.08 \\
                            & L2-norm & 87.06 & 90.94 & 85.93 \\
                            & \cellcolor{gray!15}Cosine & \cellcolor{gray!15}\textbf{89.70} & \cellcolor{gray!15}\textbf{92.52} & \cellcolor{gray!15}\textbf{88.59} \\
\midrule
\multirow{3}{*}{\rotatebox{90}{\shortstack{\textbf{8-shot}}}}  
& Random & 84.20 & 87.66 & 82.70 \\
                            & L2-norm & \textbf{84.49} & \textbf{88.40} & \textbf{83.07} \\
                            & \cellcolor{gray!15}Cosine & \cellcolor{gray!15}84.01 & \cellcolor{gray!15}88.32 & \cellcolor{gray!15}82.42 \\
\midrule
\multirow{3}{*}{\rotatebox{90}{\shortstack{\textbf{4-shot}}}}  
& Random & 72.23 & 72.58 & 65.26 \\
                            & L2-norm & 73.38 & 75.51 & \textbf{69.75} \\
                            & \cellcolor{gray!15}Cosine & \cellcolor{gray!15}\textbf{74.86} & \cellcolor{gray!15}\textbf{76.65} & \cellcolor{gray!15}68.66 \\
\bottomrule
\end{tabular}
}
\label{tab:retrieval_ablation}
\end{table}

\subsubsection{Optimization Strategy Ablation}
We investigate the impact of different optimization strategies on model performance. 
\textit{Multi-mean} optimizes the network by using the average of retrieved text features across multiple runs. 
\textit{Stochastic}, as described in Section~\ref{sec:soms}, optimizes the model using individual retrieved text instances in a stochastic manner. 
As shown in Table~\ref{tab:optimization}, the Stochastic strategy consistently outperforms Multi-mean across nearly all evaluation metrics, demonstrating the benefit of preserving semantic diversity during optimization.

\begin{table}
\centering
\caption{Performance comparison (presented in \%) of different optimization strategies in SMMO on CAMELYON under few-shot learning settings.}
\scriptsize
\renewcommand{\arraystretch}{1.3}
\resizebox{\linewidth}{!}{
\begin{tabular}{c|c|ccc}
\toprule
\textbf{} & \textbf{Optimization Strategy} & \textbf{ACC} & \textbf{AUC} & \textbf{F1 Score} \\
\midrule
\multirow{2}{*}{\rotatebox{90}{\shortstack{\textbf{16-shot}}}} 
                            & Multi-mean & 87.36 & 90.23 & 86.23 \\
                            & \cellcolor{gray!15}Stochastic & \cellcolor{gray!15}\textbf{89.70} & \cellcolor{gray!15}\textbf{92.52} & \cellcolor{gray!15}\textbf{88.59} \\
\midrule
\multirow{2}{*}{\rotatebox{90}{\shortstack{\textbf{8-shot}}}}
                            & Multi-mean & 84.08 & \textbf{89.14} & 81.70 \\
                            & \cellcolor{gray!15}Stochastic & \cellcolor{gray!15}\textbf{84.01} & \cellcolor{gray!15}88.32 & \cellcolor{gray!15}\textbf{82.42} \\
\midrule
\multirow{2}{*}{\rotatebox{90}{\shortstack{\textbf{4-shot}}}} 
                            & Multi-mean & 70.33 & 73.53 & 65.01 \\
                            & \cellcolor{gray!15}Stochastic & \cellcolor{gray!15}\textbf{74.86} & \cellcolor{gray!15}\textbf{76.65} & \cellcolor{gray!15}\textbf{68.66} \\
\bottomrule
\end{tabular}
}
\label{tab:optimization}
\end{table}

\subsubsection{Influence of LLM Choice on Knowledge Base}

\begin{table}
\centering
\caption{Performance comparison (presented in \%) of knowledge bases generated by different locally deployed LLMs under few-shot learning settings. 
}
\scriptsize
\resizebox{\linewidth}{!}{
\begin{tabular}{c|c|ccc}
\toprule
\textbf{} & \textbf{Method} & \textbf{ACC} & \textbf{AUC} & \textbf{F1 Score} \\
\midrule
\multirow{3}{*}{\rotatebox{90}{\shortstack{\textbf{16-shot}}}}
& Deepseek-R1-Distill-Qwen-7B & 87.65 & 90.93 & 86.47 \\
                            & Llama-3.2-1B & 87.17 & 90.23 & 85.95 \\
                            & \cellcolor{gray!15}Qwen2-7B & \cellcolor{gray!15}\textbf{89.70} & \cellcolor{gray!15}\textbf{92.52} & \cellcolor{gray!15}\textbf{88.59} \\
\midrule
\multirow{3}{*}{\rotatebox{90}{\shortstack{\textbf{8-shot}}}} 
& Deepseek-R1-Distill-Qwen-7B & \textbf{84.94} & \textbf{89.04} & \textbf{83.44} \\
                            & Llama-3.2-1B & 84.20 & 87.66 & 82.70 \\
                            & \cellcolor{gray!15}Qwen2-7B & \cellcolor{gray!15}84.01 & \cellcolor{gray!15}88.32 & \cellcolor{gray!15}82.42 \\
\midrule
\multirow{3}{*}{\rotatebox{90}{\shortstack{\textbf{4-shot}}}} 
& Deepseek-R1-Distill-Qwen-7B & 72.63 & 75.68 & 66.83 \\
                            & Llama-3.2-1B & 70.44 & 74.21 & 66.57 \\
                            & \cellcolor{gray!15}Qwen2-7B & \cellcolor{gray!15}\textbf{74.86} & \cellcolor{gray!15}\textbf{76.65} & \cellcolor{gray!15}\textbf{68.66} \\
\bottomrule
\end{tabular}
}
\label{tab:llm_ablation}
\end{table}

We study how the choice of large language model for knowledge base construction affects the performance of MUSE. Specifically, we generate text-based knowledge bases using three locally deployed large language models: Deepseek-R1-Distill-Qwen-7B~\cite{deepseek}, Llama-3.2-1B~\cite{llama}, and Qwen2-7B~\cite{qwen}. All models follow the knowledge generation pipeline described in Section~\ref{subsec:pipeline}. For Deepseek-R1-Distill-Qwen-7B, we disable reasoning traces to produce concise responses.

As shown in Table~\ref{tab:llm_ablation}, MUSE achieves the best performance when using the knowledge base generated by Qwen2-7B, outperforming other variants under both 4-shot and 16-shot settings. This demonstrates that the linguistic quality and semantic richness of the generated knowledge directly influence downstream few-shot classification accuracy. The results highlight the importance of selecting capable large language models when constructing knowledge bases for medical vision-language learning.

\subsection{Visualization Analysis}

\begin{figure} 
	\centering
	\includegraphics[trim=0 0 0 0,clip,scale=0.142]{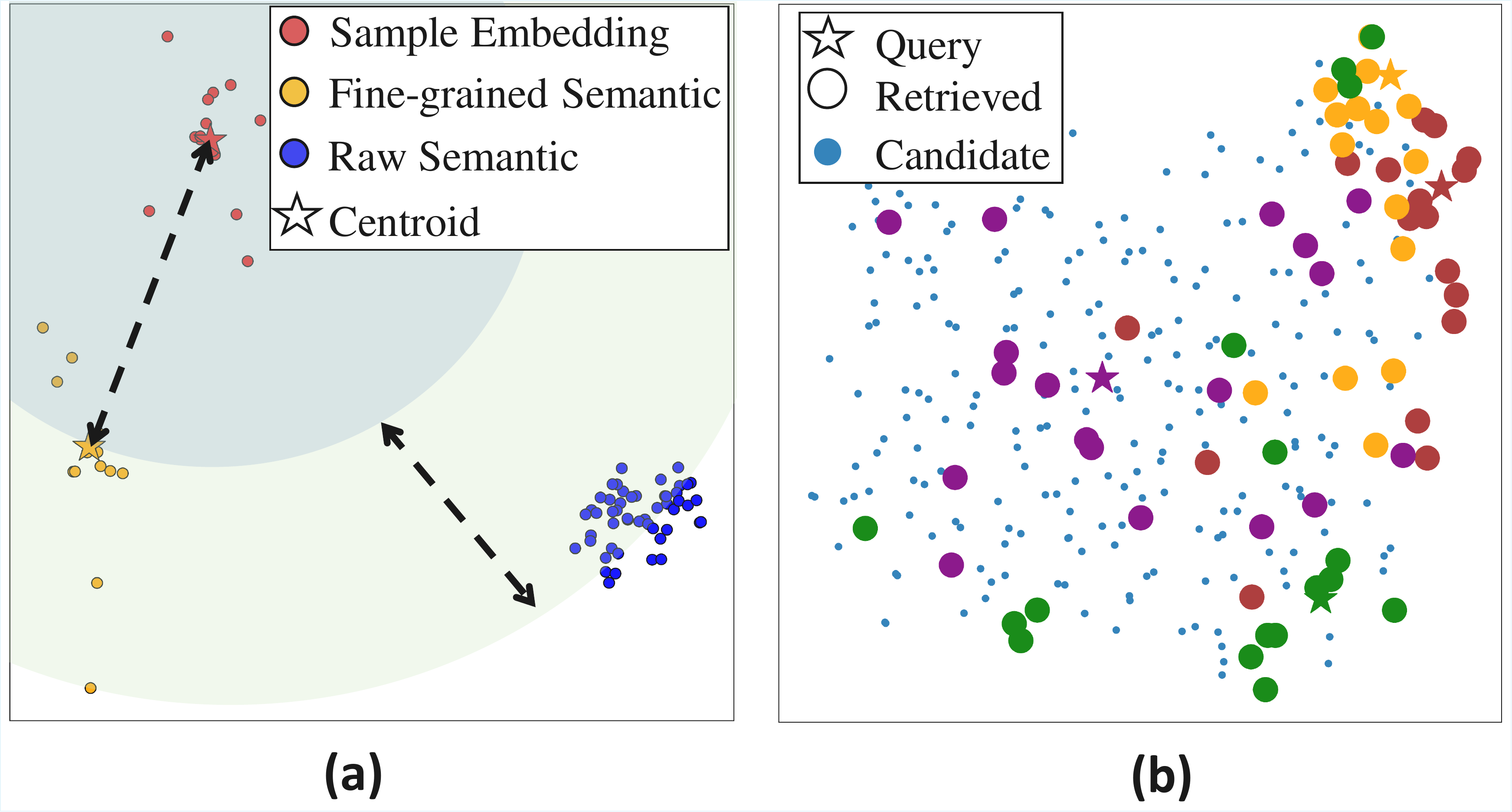} 
	\caption{The visualization analysis of fine-grained semantic priors via SFSE and sample-based retrieval. (a) Visualization of fine-grained semantic in sample embedding space. (b) Comparison of text features retrieved by different samples. (Different colors represent different samples.)
}
	\label{visualization}
\end{figure}

To investigate the alignment between fine-grained semantics and sample embeddings, we jointly map the raw semantics and fine-grained semantic priors into the unified space. The visualization in Figure~\ref{visualization} (a) shows that, in the latent space, the fine-grained semantic priors refined by SFSE are closer to the sample embeddings than the raw semantics. Moreover, we compare the behavior of text feature retrieval from the knowledge base across different samples, with visualization shown in Figure~\ref{visualization} (b). This demonstrates that our method is capable of retrieving diverse and semantically rich textual features from the knowledge base, with retrieved features exhibiting distinctiveness across different samples. Additional details on visualizations and the properties of MUSE are provided in the Supplementary Material.

%% file: sec/5_conclusion.tex
\section{Conclusion}

In this study, we propose MUSE, a stochastic multi-view semantic enhancement framework for few-shot whole slide image classification. We design the Sample-wise Fine-grained Semantic Enhancement module, which leverages a Mixture-of-Experts architecture to decompose category semantics and achieve fine-grained alignment between visual and textual representations. Furthermore, we introduce the Stochastic Multi-view Model Optimization module, which leverages the large language models to construct a multi-view textual knowledge base for each pathological category and exploits the diversity of generated semantics through retrieval and stochastic optimization. Across multiple standard pathology datasets, our framework consistently demonstrates superior performance in few-shot scenarios. This shows that enhanced semantic precise and enriched semantic diversity can effectively improve the model’s ability to understand and generalize complex pathological features, thereby paving a new technical pathway for computational pathology analysis.

%% file: sec/X_suppl.tex
\clearpage
\setcounter{page}{1}
\maketitlesupplementary

\section{Knowledge Base Generation Pipeline}
\subsection{Concept Decomposition}
To consider the visual characteristics description of pathological images from multiple perspectives, we utilize the prompt to query GPT-4 \cite{gpt4} and summarize the diagnostic-related visual characteristics into four aspects: cellular morphology, tissue architecture, color-staining characteristics, and spatial-texture patterns. The prompt we used is: \textit{You are an experienced pathologist. For the \{dataset\}, starting from visual features, what are four aspects that can be used for diagnosis? Summarize into four aspects}.

\subsection{Example Generation}
For each aspect, we query GPT-4 to generate 10 specific examples for each category, and these examples could guide the subsequent LLM to generate more accurate responses. The prompt we used in generating examples is: \textit{From \{perspective\}, what are the characteristics of WSI images for \{class name\}? Please list 10 examples.} Specific examples for CAMELYON are presented in Section \ref{examples}.

\subsection{Multi-view Category Knowledge Base Construction}
For each category in the dataset, we randomly select one example from each aspect
as the description for that perspective, and combine these examples into a query prompt
to query subsequent LLM. The query prompt is: \textit{To supplement and systematize the description of \{class name\} in high-resolution imaging from four perspectives: \{example from each perspective\}. Please summarize it into a single paragraph.} In this prompt, \textit{\{example from each perspective\}} is the example we selected for each perspective. For each category in every dataset, we generated 300 pathological descriptions using the above-mentioned generation procedure, forming the multi-view category knowledge base. We sampled one example per category from the generated knowledge base for each dataset, as shown in Table~\ref{tab:knowledge_base}.

\begin{table*}[t]
\centering
\caption{Examples of sampled descriptions from the multi-view category knowledge base for each category within each dataset.
}
\scriptsize
\renewcommand{\arraystretch}{1.4}
\begin{tabularx}{\textwidth}{>{\centering\hsize=0.2\hsize}X|>{\centering\hsize=0.2\hsize}X|>{\hsize=2.6\hsize}X}
\cmidrule(lr){1-3}
\multicolumn{1}{c}{\textbf{}} & 
\multicolumn{1}{|c|}{\textbf{Category}} & 
\multicolumn{1}{c}{\textbf{Description}} \\
\cmidrule(lr){1-3}

\multirow{7}{*}{\rotatebox{90}{\shortstack{\textbf{CAMELYON}}}}
& Normal 
& In high-resolution imaging, a normal lymph node exhibits uniform nuclear staining with homogeneous chromatin that lacks coarse granularity or heterogeneity. The vascular distribution is regular, featuring well-organized capillaries and small vessels without any signs of abnormal dilation or neovascularization. There are no non-physiological speckles, such as pigment deposits, debris, or staining artifacts, and the boundaries between different tissue regions show smooth transitions without abrupt changes or protrusions. \\

\cmidrule(lr){2-3}
& Tumor 
&  In high-resolution imaging of a metastatic lymph node, cellular crowding is evident with densely packed cells often forming clusters or sheets, indicative of significant tumor burden. Concurrently, irregular or distorted vascular structures are observed, showing signs of compression, dilation, or irregular branching of blood vessels, suggesting compromised microvascular integrity. Additionally, increased nuclear basophilia is noted, characterized by darker purple nuclei due to elevated DNA content in the tumor cells, reflecting their proliferative activity. Lastly, diffuse infiltration patterns are present, where tumor cells spread along existing tissue frameworks, displacing normal lymphoid architecture and disrupting the orderly arrangement of lymphocytes and other immune cells.\\
\cmidrule(lr){1-3}

\multirow{11}{*}{\rotatebox{90}{\shortstack{\textbf{TCGA-NSCLC}}}}
& LUAD 
& Lung adenocarcinoma exhibits a complex morphology that can be systematically described through high-resolution imaging. Notably, some cells display vacuolated or mucinous cytoplasmic changes, indicating alterations in cellular metabolism and possibly reflecting the presence of secretory activity. Additionally, the tumor often features fibrovascular cores within its papillary structures, which provide structural support and potentially contribute to the tumor's invasive capacity. Angiogenesis is another key feature, characterized by the presence of numerous, thin-walled blood vessels in the tumor stroma, which are crucial for tumor growth and survival. Lastly, there is significant heterogeneity in mucin content, with certain regions producing abundant mucin, while others show little or no production, suggesting diverse microenvironments and potential for different clinical behaviors. \\

\cmidrule(lr){2-3}
& LUSC 
& In high-resolution imaging of lung squamous cell carcinoma, key features can be systematically described from four perspectives: Intercellular bridges manifest as visible desmosomal connections between tumor cells, indicating cellular cohesion. Polygonal tumor cells are characterized by well-defined, often angular shapes, reflecting the distinctive morphology of these neoplastic cells. The tumor-stroma interface exhibits heterogeneity, with irregular boundaries that highlight the complex interaction between the malignant cells and their microenvironment. Lastly, local variation in mitotic activity is evident, with hotspots of proliferating cells indicating areas of increased cellular division and potential for aggressive growth. \\ 
\cmidrule(lr){1-3}

\multirow{7}{*}{\rotatebox{90}{\shortstack{\textbf{TCGA-BRCA}}}}
& IDC 
& Invasive ductal carcinoma exhibits high cellular density with cells closely packed and minimal intervening stroma, leading to a crowded appearance. Ducts often show lumen formation with cellular debris accumulating within them. Surrounding these areas, peritumoral fibrosis is characterized by thickened fibrotic bands that encircle the invasive tumor regions. Additionally, desmoplastic stroma is evident as dense fibrous tissue frequently surrounds the tumor clusters, contributing to a robust and complex microenvironment. \\

\cmidrule(lr){2-3}
& ILC 
& Invasive lobular carcinoma (ILC) can be characterized through high-resolution imaging from four distinct perspectives: In single-file arrangement, tumor cells often infiltrate in linear chains between collagen fibers, creating a linear or string-of-beads appearance. In diffuse infiltration, tumor cells spread in a more scattered, non-nodular manner, leading to a less organized and more widespread distribution. There is typically minimal fibroblast activation with fewer myofibroblasts compared to invasive ductal carcinoma (IDC), resulting in a less fibrotic stroma. Additionally, a targetoid pattern may be observed where tumor cells wrap around normal ductal structures, creating a concentric or target-like appearance around these structures. \\
\cmidrule(lr){1-3}
\end{tabularx}
\label{tab:knowledge_base}
\end{table*}

\section{Implementation Details}
In the DSR module, we set 8 query matrices as experts for fine-grained semantic processing, and a router network is employed to score these experts. The top-2 experts with the highest scores are selected for further processing. In the SVTI module, we retain only the top-20\% patches with the highest attention scores for each head. In the SMMO component, we use the fine-grained semantic prior to retrieve the top-20 most similar text features from the knowledge base for subsequent semantic diversity optimization and enhancement. An Adam optimizer \cite{adam} with learning rate of $1\times 10^{-4}$ and weight decay of $1\times10^{-5}$ is used for the model training. All the models are trained for 200 epochs with an early-stopping strategy. The batch size is set to 1. All the experiments are conducted with a single NVIDIA RTX 3090 GPU.

\section{Training and Inference Strategy}

During training, for each WSI bag, we generate fine-grained semantic priors using the SFSE module as follows: the DSR component first decomposes the raw semantic information into specialized representations, which are then utilized as fine-grained cues in the SVTI module. These fine-grained semantic priors serve as queries to retrieve relevant textual description features from the category-specific knowledge base. A stochastic optimization strategy is further adopted to leverage the retrieved semantic diversity. The detailed training procedure is outlined in Algorithm~\ref{alg:training}.

During inference, due to the absence of category priors, we input both the class names of the dataset and the test WSI into the SFSE module as well as the classifier to obtain the final prediction. The detailed algorithm is presented in Algorithm~\ref{alg:inference}.

\begin{algorithm}[!h]
    \caption{Training Strategy}
    \label{alg:training}
    \renewcommand{\algorithmicrequire}{\textbf{Input:}}
    \renewcommand{\algorithmicensure}{\textbf{Output:}}
    
    \begin{algorithmic}[1]
        \REQUIRE Training set of WSI bags $\{B^{l}\}_{l=1}^{L}$ with labels $\{Y^{l}\}_{l=1}^{L}$, category text features $D$,
        number of experts $R$, number of selected experts $k$, top-patch ratio $r\%$, number of retrieved texts $m$, maximum epoch $E_{max}$.

        \STATE Initialize parameters of module DSR, SVTI, and MLP
        \FOR{epoch $e=1$ to $E_{max}$}
            \FOR{each WSI bag $B^{l}$ in training set}
                \STATE // Fine-grained semantic prior $f$ is obtained through the DSR and SVTI modules
                \STATE $Q=\text{DSR}(D;k,R)$
                \STATE $f=\text{SVTI}(B^{l}, Q; r\%)$
                \STATE $z=\text{MLP}(f)$
                \STATE // Retrieve relevant semantics from the category-related text knowledge base
                \STATE $T_m$: The randomly shuffled top $m$ retrieved text features from $\mathcal{B}_{Y^{l}}$
                \FOR{$t \in T_m$}
                    \STATE $\{Q^{t}_{j}\}_{j=1}^{k} = \text{DSR}(t;k,R)$
                    \STATE $f^{t}_{\text{aux}} = \text{SVTI}(B^{l},\{Q^{t}_{j}\}_{j=1}^{k};r\%)$
                    \STATE $z^{t}_{\text{aux}} = \text{MLP}(f^{t}_{\text{aux}})$
                    \STATE $z^{t}_{\text{final}} = \frac{z + z^{t}_{\text{aux}}}{2}$
                    \STATE $\mathcal{L}^{t} = \text{CrossEntropyLoss}(z^{t}_{\text{final}}, Y^{l})$
                    \STATE update \text{DSR}, \text{SVTI}, \text{MLP} by back-propagation via Adam W optimizer
                \ENDFOR
            \ENDFOR
        \ENDFOR
        \STATE return Optimal modules $\text{DSR}^{*}$, $\text{SVTI}^{*}$, and $\text{MLP}^{*}$
    \end{algorithmic}
\end{algorithm}

\begin{algorithm}[!h]
    \caption{Inference Strategy}
    \label{alg:inference}
    \renewcommand{\algorithmicrequire}{\textbf{Input:}}
    \renewcommand{\algorithmicensure}{\textbf{Output:}}
    
    \begin{algorithmic}[1]
        \REQUIRE Test WSI bag $B$ with label $Y$, category text features $D$,optimal modules $\text{DSR}^{*}$, $\text{SVTI}^{*}$, and $\text{MLP}^{*}$.
        
        
        
        \STATE // Fine-grained semantic prior $f$ is obtained through the DSR and SVTI modules
        \STATE $Q = \text{DSR}(D;k,R)$
        \STATE $f = \text{SVTI}(B,Q;r\%)$ 
        \STATE $z = \text{MLP}(f)$
        \STATE // Label prediction via argmax over logit
        \STATE $\hat{y}=\text{argmax}(z)$
        \STATE return predicted label $\hat{y}$

    \end{algorithmic}
\end{algorithm}

\section{Details of Visualization}
In the visualization of fine-grained semantics in the sample embedding space, we sample some textual description features from the category knowledge base to serve as raw semantics. We sample WSIs from the corresponding category and obtain the sample embeddings by averaging all patch features within the bags. Subsequently, we employ the SFSE module to perform fine-grained semantic modeling on these WSI bags, obtaining fine-grained semantic priors. We then use t-SNE \cite{tsne} visualization to project sample embeddings, raw semantics, and fine-grained semantic priors into a shared embedding space for visual comparison.

In the comparison of text features retrieved by different samples, we sample several WSIs and obtain the corresponding fine-grained semantic priors using the SFSE module. These priors are then used as queries to retrieve relevant text features from the category knowledge base. Finally, we visualize the query, the candidate text features in the knowledge base, and the retrieved text features in a shared embedding space using t-SNE.

\section{Additional Quantitative Experiments}
\subsection{Effects of Pathology Foundation Model Encoders}
To evaluate the effectiveness of MUSE with different pathology foundation models (FM) serving as feature extractors, we compared four distinct pathological visual-language foundation models, i.e., PLIP \cite{plip}, MUSK \cite{musk}, Prov-GigaPath \cite{prov-giga}, and CONCH \cite{conch}, under 16-shot setting across ACC, AUC, and F1 Score metrics. The results are presented in Table~\ref{tab:fm_ablation}. The results demonstrate that CONCH outperforms other FMs across all evaluation metrics. This stems from CONCH's explicit adoption of a CLIP-style dual-tower architecture \cite{clip}, which enables effective understanding of pathological text semantics. In contrast, Prov-GigaPath is primarily designed as an image-centric model with self-supervised learning \cite{mae}; although it incorporates textual information for fine-tuning, its core remains a unimodal image encoder, thereby limiting its capability in text comprehension.

\begin{table}
\centering
\caption{Performance comparison (presented in \%) of different foundation models (PLIP, MUSK, Prov-GigaPath, and CONCH) on CAMELYON under 16-shot settings.}
\scriptsize
\resizebox{\linewidth}{!}{
\begin{tabular}{c|ccc}
\toprule
\textbf{FM} & \textbf{ACC} & \textbf{AUC} & \textbf{F1 Score} \\
\midrule
PLIP \cite{plip}         & 72.82 & 73.47 & 67.24 \\
MUSK \cite{musk} & 86.72 & 88.28 & 85.18 \\
Prov-GigaPath \cite{prov-giga} & 87.32 & 89.73 & 84.26 \\
\cellcolor{gray!15}CONCH \cite{conch} & \cellcolor{gray!15}\textbf{89.70} & \cellcolor{gray!15}\textbf{92.52} & \cellcolor{gray!15}\textbf{88.59} \\

\bottomrule
\end{tabular}
}
\label{tab:fm_ablation}
\end{table}

\subsection{Ablation on the Number of Retrieved Texts}
We conduct ablation experiments to investigate the performance across different values of the number of retrieved texts in the SMMO component. The value of this hyperparameter is set to 10, 20, and 80, respectively. The experimental results are presented in Table~\ref{tab:retrieval_text}. Under the 16-shot setting, the optimal performance is achieved when the hyperparameter is set to 20. For the 8-shot and 4-shot settings, the best results are obtained with a value of 80. This indicates that in scenarios with sparse visual signals, increasing semantic diversity contributes to improved model generalization.

\begin{table}
\centering
\caption{Performance comparison (presented in \%) of different number of retrieved texts on CAMELYON under few-shot learning settings.}
\scriptsize
\renewcommand{\arraystretch}{1.3}
\resizebox{\linewidth}{!}{
\begin{tabular}{c|c|ccc}
\toprule
\textbf{} & \textbf{number of retrieved texts} & \textbf{ACC} & \textbf{AUC} & \textbf{F1 Score} \\
\midrule
\multirow{3}{*}{\rotatebox{90}{\shortstack{\textbf{16-shot}}}} 
                            & 10 & 87.88 & 91.33 & 86.75 \\
                            & 20 & \textbf{89.70} & \textbf{92.52} & \textbf{88.59} \\
                            & 80 & 88.43 & 91.63 & 87.26 \\
                            
\midrule
\multirow{3}{*}{\rotatebox{90}{\shortstack{\textbf{8-shot}}}}
                            & 10 & 83.64 & 86.44 & 80.79 \\
                            & 20 & 84.01 & 88.32 & 82.42 \\
                            & 80 & \textbf{84.49} & \textbf{88.40} & \textbf{83.07} \\
\midrule
\multirow{3}{*}{\rotatebox{90}{\shortstack{\textbf{4-shot}}}} 
                            & 10 & 69.14 & 71.41 & 64.01 \\
                            & 20 & 74.86 & 76.65 & 68.66 \\
                            & 80 & \textbf{77.32} & \textbf{77.95} & \textbf{72.53} \\
\bottomrule
\end{tabular}
}
\label{tab:retrieval_text}
\end{table}

\subsection{Sensitivity to the Number of Experts}
To investigate the sensitivity of model performance to the number of experts in the DSR module, we conducted experiments with the number of experts set to 4, 8, and 16, while consistently selecting the top-2 experts based on gating scores. As shown in Table~\ref{tab:num_expert}, the experimental results demonstrate that 8-experts achieves the best performance under the 16-shot setting, while 4-experts yield the optimal results in the 4-shot setting.

\begin{table}
\centering
\caption{Performance comparison (presented in \%) of different number of experts on CAMELYON under few-shot learning settings.}
\scriptsize
\renewcommand{\arraystretch}{1.3}
\resizebox{\linewidth}{!}{
\begin{tabular}{c|c|ccc}
\toprule
\textbf{} & \textbf{number of experts} & \textbf{ACC} & \textbf{AUC} & \textbf{F1 Score} \\
\midrule
\multirow{3}{*}{\rotatebox{90}{\shortstack{\textbf{16-shot}}}} 
                            & 4 & 87.91 & 90.38 & 86.64 \\
                            & 8 & \textbf{89.70} & \textbf{92.52} & \textbf{88.59} \\
                            & 16 & 87.32 & 90.16 & 86.18 \\
                            
\midrule
\multirow{3}{*}{\rotatebox{90}{\shortstack{\textbf{8-shot}}}}
                            & 4 & 82.88 & 87.49 & 81.52 \\
                            & 8 & 84.01 & \textbf{88.32} & \textbf{82.42} \\
                            & 16 & \textbf{84.31} & 87.60 & 81.94 \\
\midrule
\multirow{3}{*}{\rotatebox{90}{\shortstack{\textbf{4-shot}}}} 
                            & 4 & \textbf{75.31} & \textbf{77.43} & \textbf{68.90} \\
                            & 8 & 74.86 & 76.65 & 68.66 \\
                            & 16 & 72.49 & 75.01 & 68.27 \\
\bottomrule
\end{tabular}
}
\label{tab:num_expert}
\end{table}

\subsection{Ablation on the Top-Patch Ratio}
We compare different top-patch ratios in DSR module to investigate the impact of this hyperparameter on model performance. We set the top-patch ratios to 10\%, 20\%, 50\%, and 100\% (retaining all patches), and conducted experiments on CAMELYON under 16-shot, 8-shot, and 4-shot settings. The results are presented in Table~\ref{tab:patch_ratio}. The results show that interacting with a subset of patches at an appropriate ratio yields better performance than retaining all patches. This indicates that adopting an appropriate top-patch ratio can effectively suppress the substantial redundant noise inherent in patches, thereby enhancing the discriminability of the interacted patch representations.

\begin{table}
\centering
\caption{Performance comparison (presented in \%) of different top-patch ratio on CAMELYON under few-shot learning settings.}
\scriptsize
\renewcommand{\arraystretch}{1.2}
\resizebox{\linewidth}{!}{
\begin{tabular}{c|c|ccc}
\toprule
\textbf{} & \textbf{Ratio} & \textbf{ACC} & \textbf{AUC} & \textbf{F1 Score} \\
\midrule
\multirow{4}{*}{\rotatebox{90}{\textbf{16-shot}}}
                            & 10\% & 88.92 & 91.92 & 87.83 \\
                            & 20\% & \textbf{89.70} & \textbf{92.52} & \textbf{88.59} \\
                            & 50\% & 88.51 & 91.66 & 87.40 \\
                            & 100\% & 88.51 & 91.62 & 87.53 \\
                            
\midrule
\multirow{4}{*}{\rotatebox{90}{\textbf{8-shot}}}
                            & 10\% & 85.65 & 88.58 & 83.65 \\
                            & 20\% & 84.01 & 88.32 & 82.42 \\
                            & 50\% & \textbf{85.57} & \textbf{90.82} & \textbf{84.28} \\
                            & 100\% & 84.34 & 86.98 & 82.54 \\
\midrule
\multirow{4}{*}{\rotatebox{90}{\textbf{4-shot}}}
                            & 10\% & 71.15 & 73.25 & 67.26 \\
                            & 20\% & \textbf{74.86} & \textbf{76.65} & \textbf{68.66} \\
                            & 50\% & 70.89 & 72.72 & 62.34 \\
                             & 100\% & 69.10 & 70.17 & 62.97 \\
\bottomrule
\end{tabular}
}
\label{tab:patch_ratio}
\end{table}

\section{Additional Visualization}
\subsection{Visualization of Attended Patch Features by Experts}

\begin{figure} 
	\centering
	\includegraphics[trim=0 0 0 0,clip,scale=0.13]{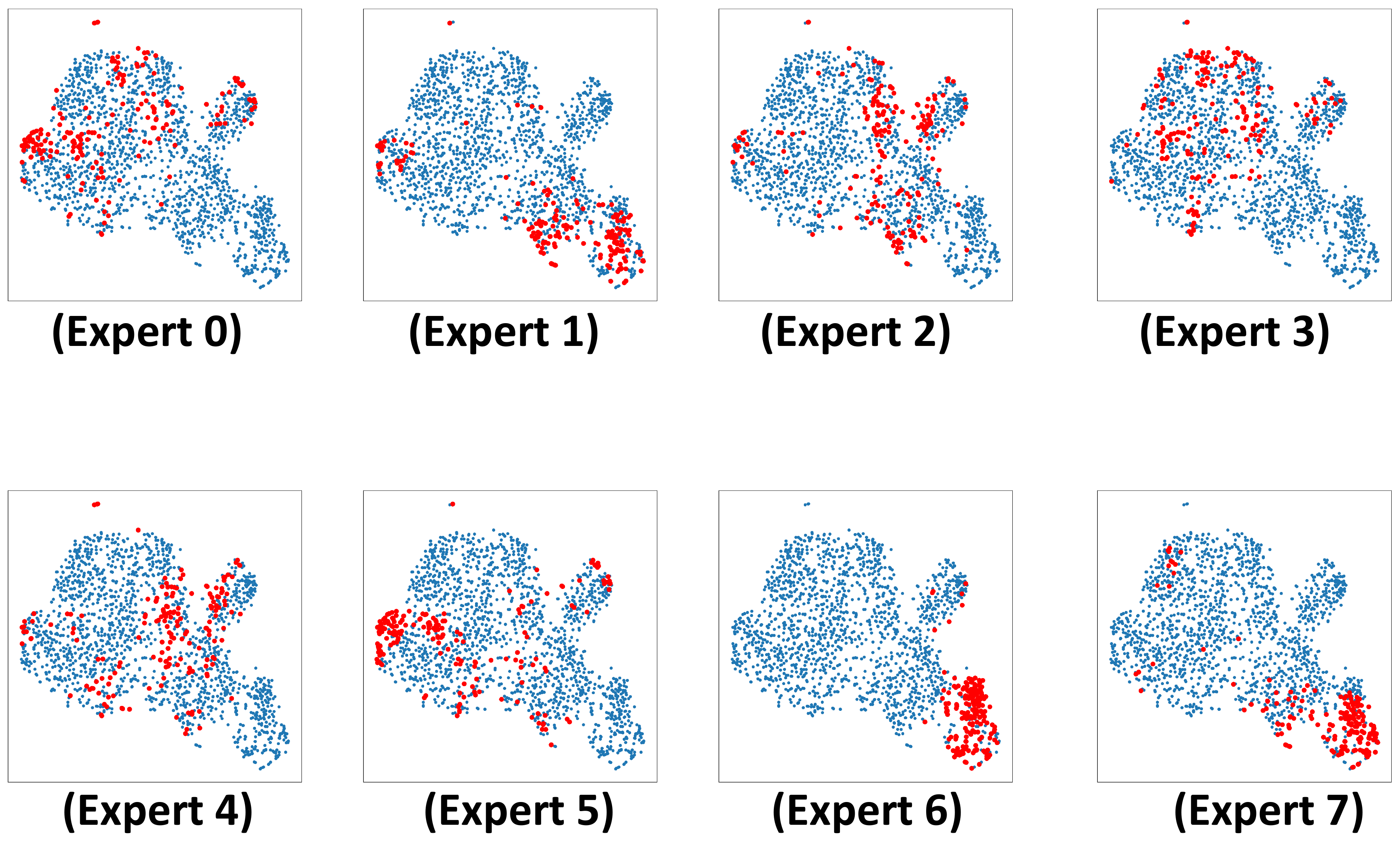} 
	\caption{Visualization of the regions attended by individual experts.}
	\label{fig:8experts}
\end{figure}

To intuitively analyze the relationship between the semantic knowledge decomposed by each expert and the patch features, we compute the similarity between the semantic information processed by each expert and the key projections derived from the patch features. We then select the top-20\% of patches based on their similarity scores and highlight the value projections obtained from these patches. As shown in Figure~\ref{fig:8experts}, this visualization reveals the specific regions attended by each expert. The visualization results demonstrate that different experts attend to distinct aspects of the patch features, indicating specialization in capturing diverse semantic patterns.

\begin{figure} 
	\centering
	\includegraphics[trim=0 0 0 0,clip,scale=0.13]{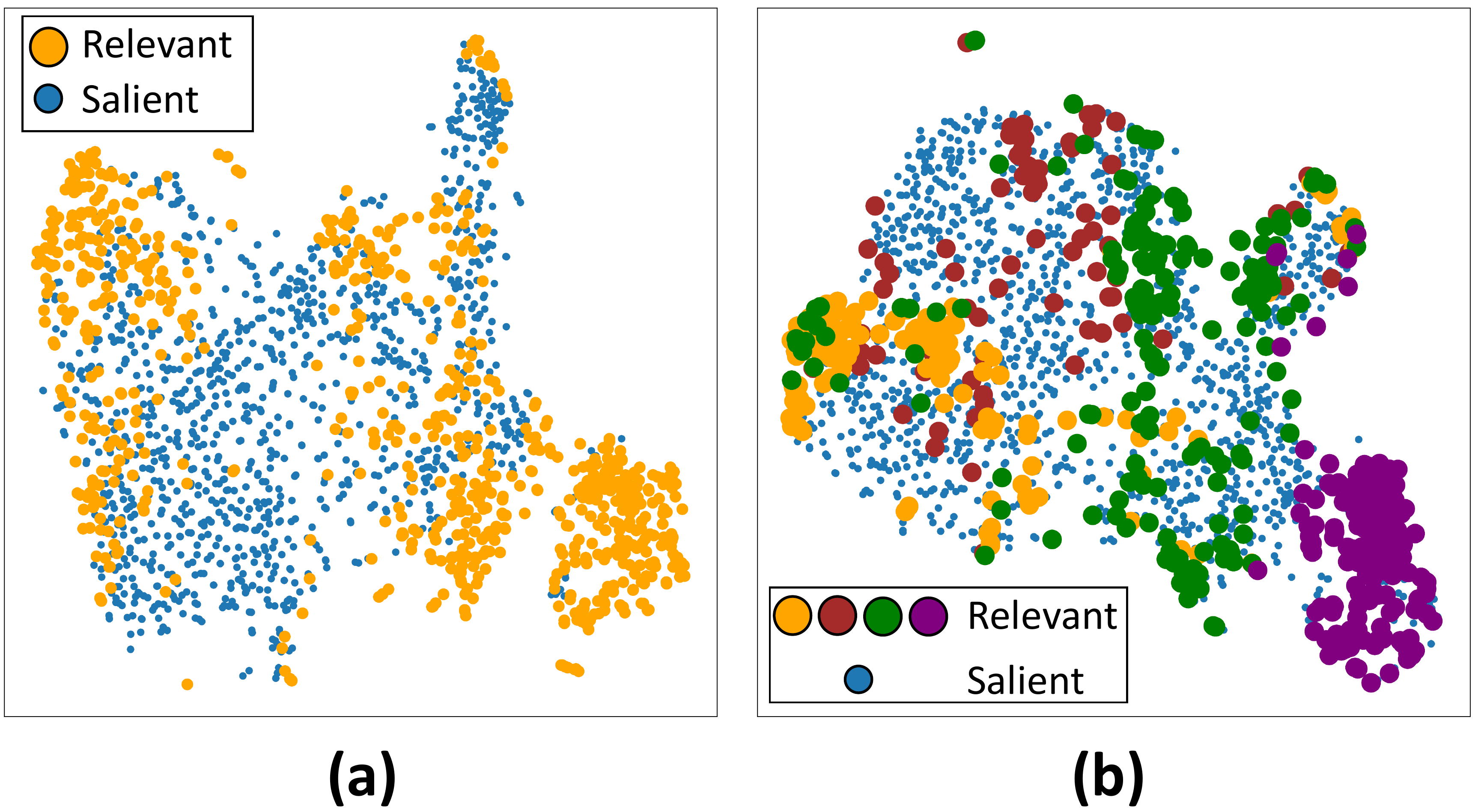} 
	\caption{The visualization analysis of semantic refinement via DSR. (a) Visual regions associated with text mapping through the cross-attention mechanism. (b) Relevant visual regions after mapping via DSR.}
	\label{fig:relevant}
\end{figure}

To visually demonstrate the semantic refinement decomposition achieved by DSR and its corresponding alignment with visual features, we provide the visualization analysis as shown in Figure~\ref{fig:relevant}. The comparisons demonstrate that, compared to cross-attention, DSR enables fine-grained semantic modeling and focuses on more concentrated regions.

\subsection{Semantic Retrieval Results for one WSI Instance}
\begin{figure} 
	\centering
	\includegraphics[trim=0 0 0 0,clip,scale=0.098]{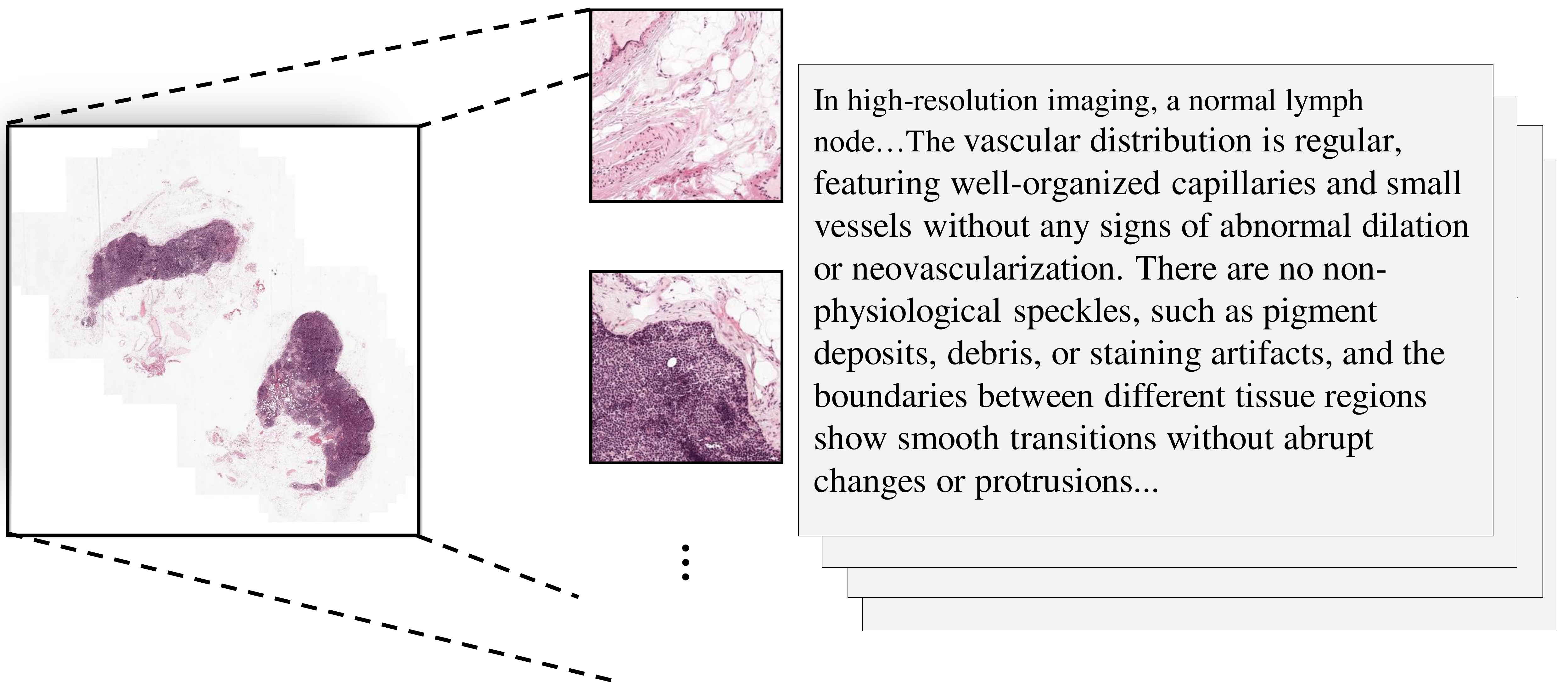} 
	\caption{Retrieved textual descriptions from the knowledge base given a WSI query.}
	\label{fig:example}
\end{figure}

We select WSI normal\_001 in CAMELYON 16 as the query to retrieve the corresponding textual descriptions from the knowledge base, as shown in Figure~\ref{fig:example}. This indicates that the retrieved textual descriptions are consistent with the image characteristics of the WSI.

\section{Limitation}
Although the DSR module aligns fine-grained semantics with patch features by leveraging the Mixture-of-Experts for refined semantic modeling, performance improvements on the benchmark alone, along with qualitative visualizations, are insufficient to fully validate its interpretability. Specifically, we do not employ stronger inductive biases to enforce specialization among the experts, but rather rely solely on data-driven adaptation to guide their learning. This limitation highlights a promising direction for future research: incorporating informative priors to constrain the learning process of experts, thereby enhancing the interpretability of semantic perception.

\section{Examples generated by LLM for CAMELYON}\label{examples}
We utilize GPT-4 to generate 10 examples of descriptions for each category in CAMELYON, covering four aspects (Cellular morphology, Tissue architecture, Color-staining, and Spatial-texture pattern), as shown on the following pages.

\begin{figure*} 
	\centering
	\includegraphics[trim=0 0 0 0,clip,scale=0.3]{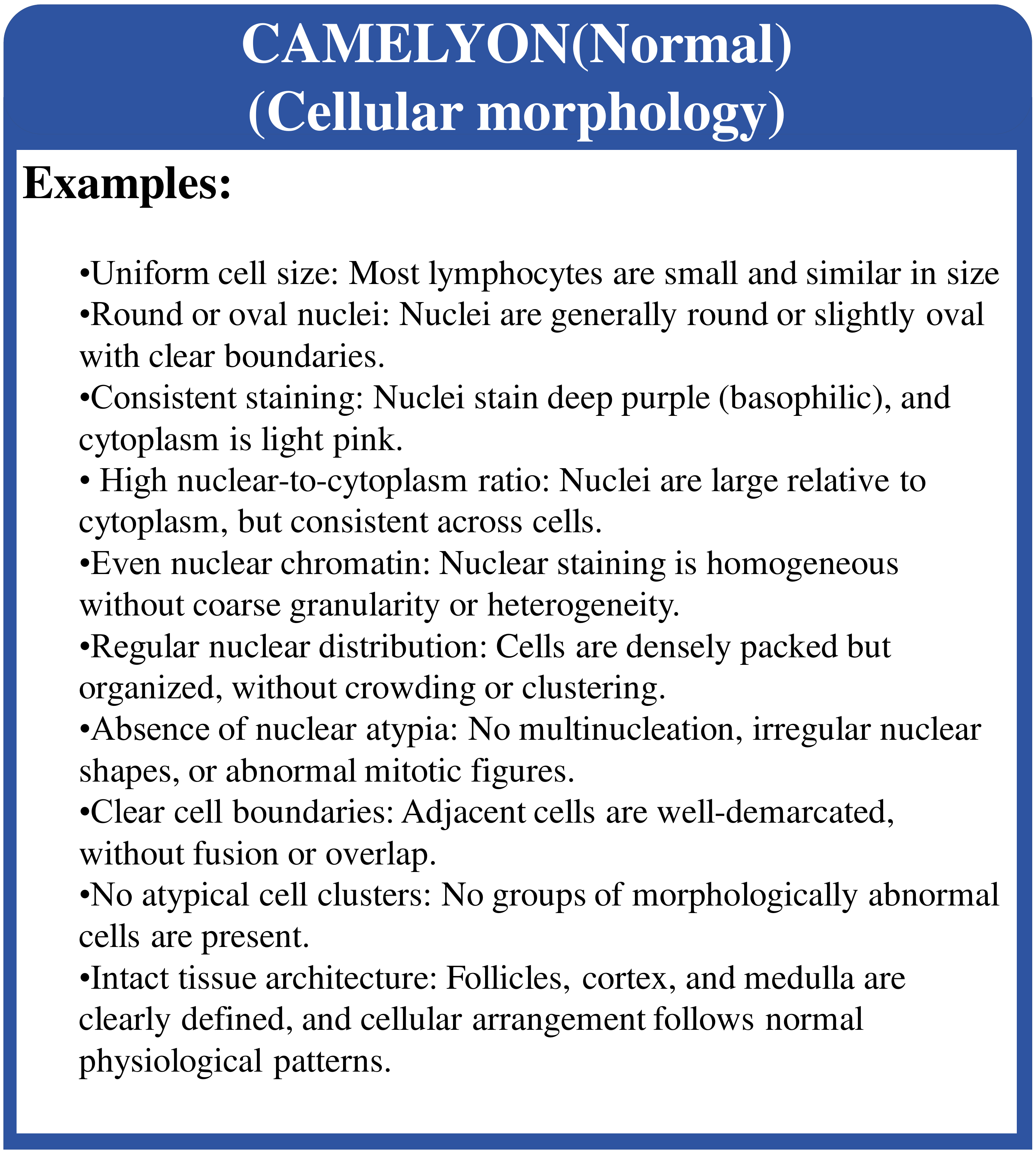} 
	\label{normal1}
\end{figure*}

\begin{figure*} 
	\centering
	\includegraphics[trim=0 0 0 0,clip,scale=0.3]{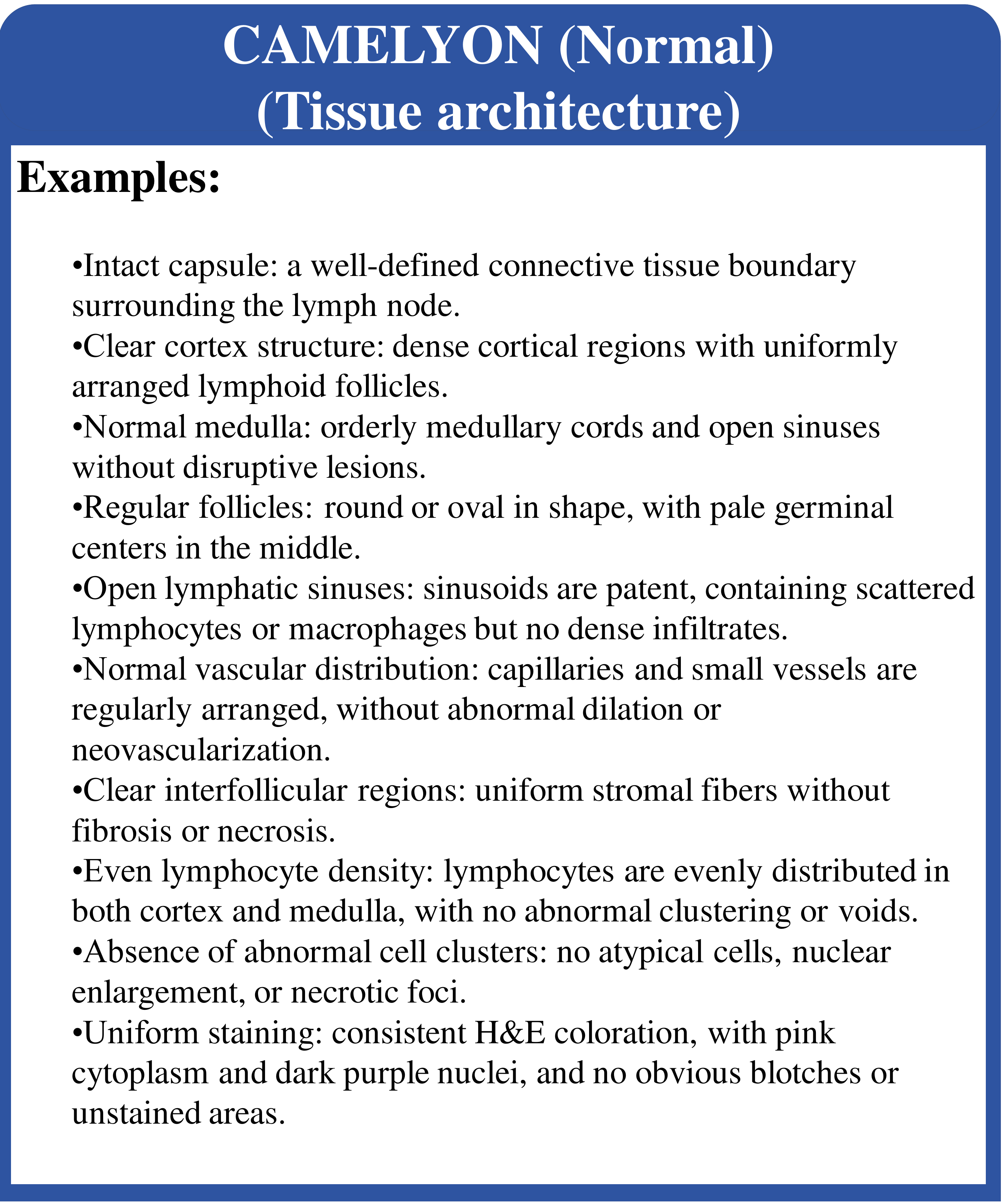} 
	\label{normal2}
\end{figure*}

\begin{figure*} 
	\centering
	\includegraphics[trim=0 0 0 0,clip,scale=0.3]{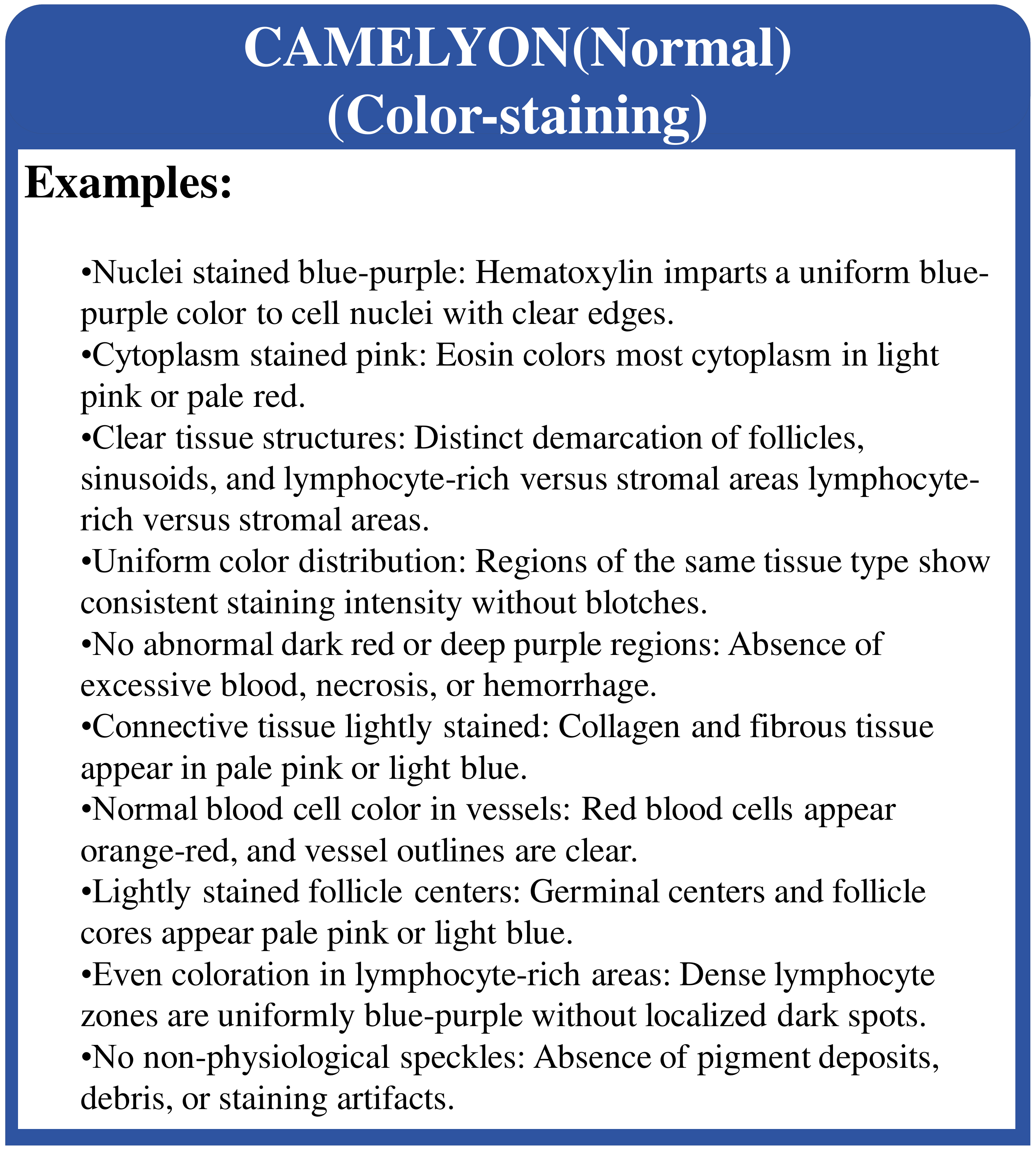} 
	\label{normal3}
\end{figure*}

\begin{figure*} 
	\centering
	\includegraphics[trim=0 0 0 0,clip,scale=0.3]{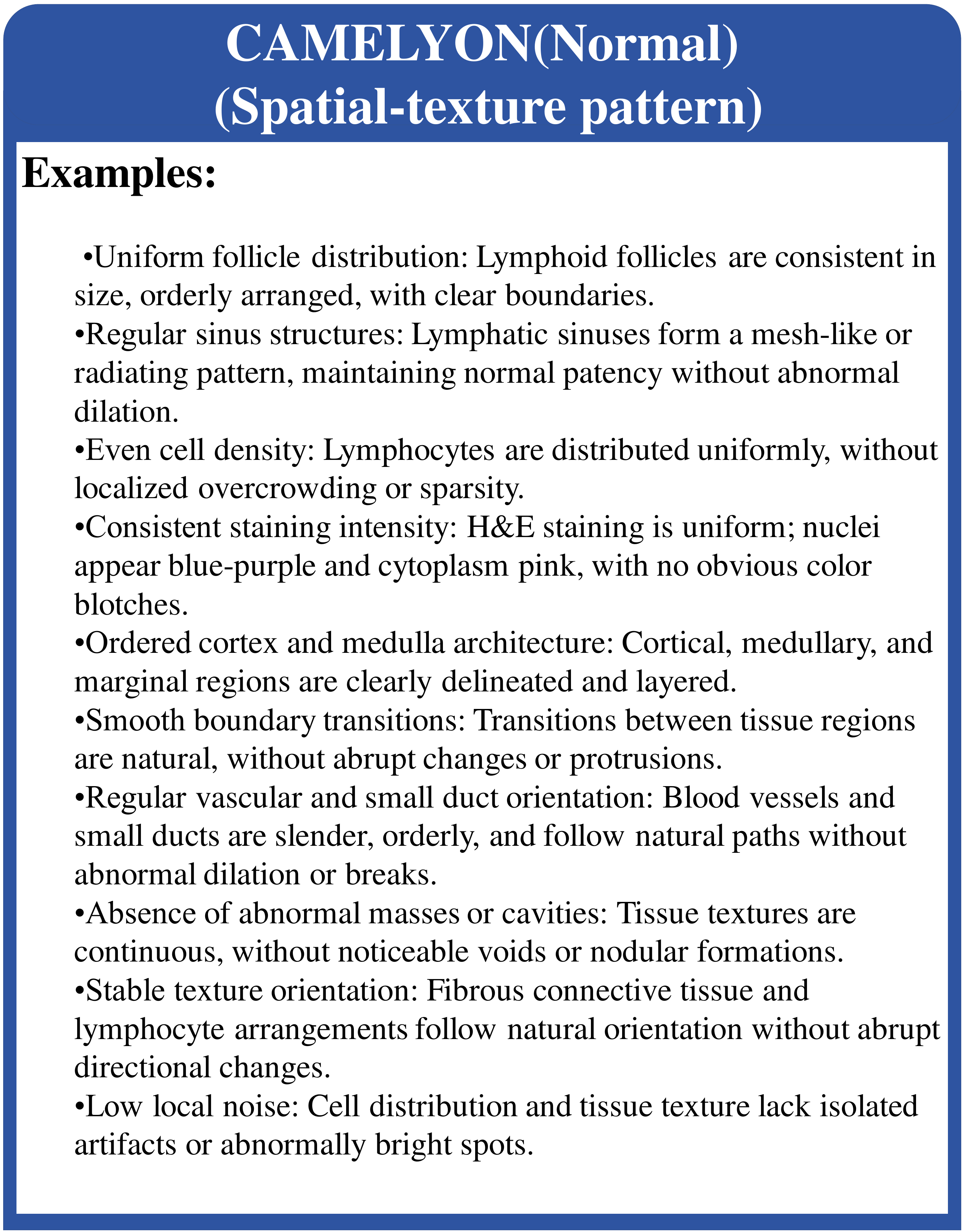} 
	\label{normal4}
\end{figure*}

\begin{figure*} 
	\centering
	\includegraphics[trim=0 0 0 0,clip,scale=0.3]{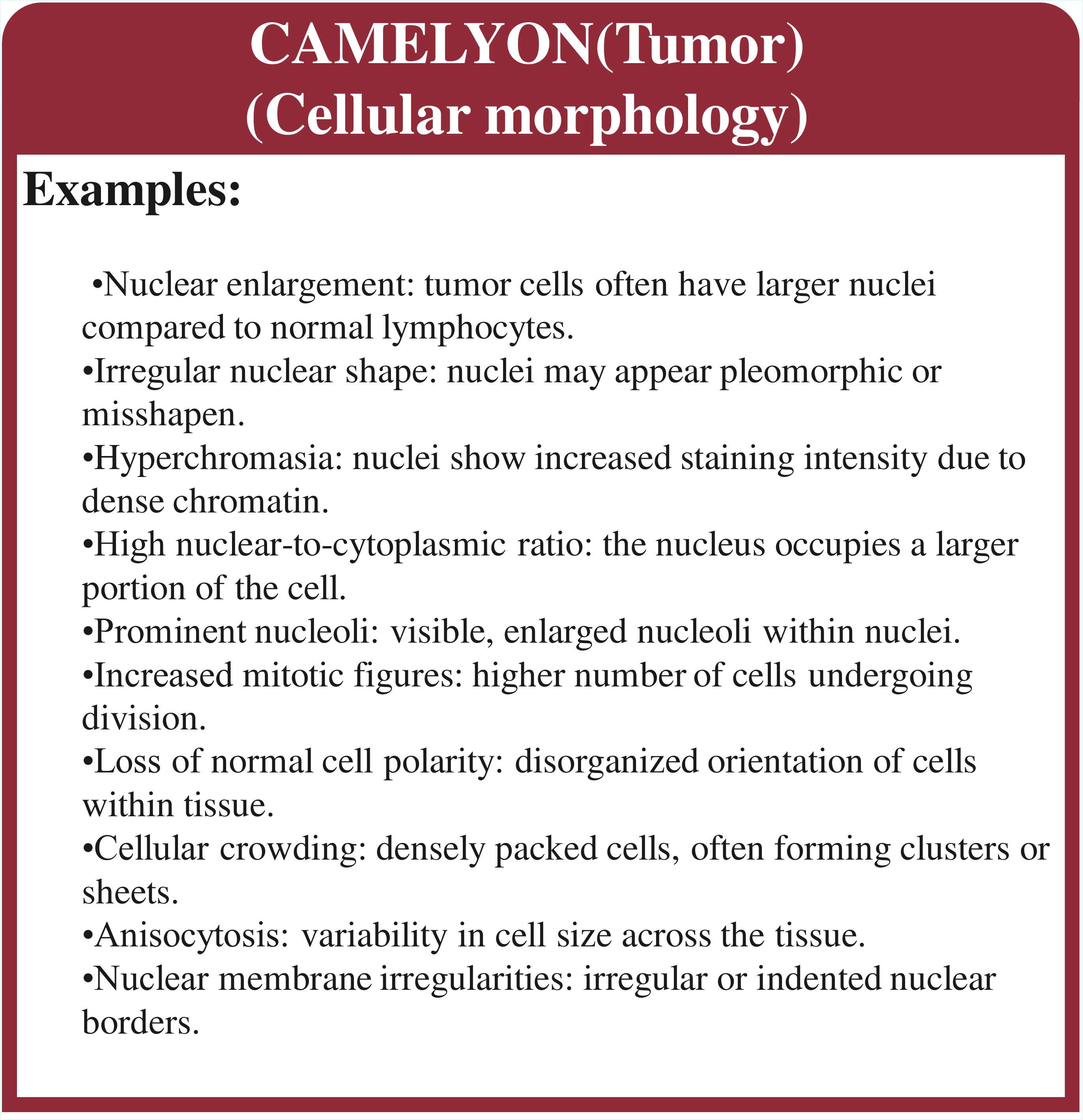} 
	\label{tumor1}
\end{figure*}

\begin{figure*} 
	\centering
	\includegraphics[trim=0 0 0 0,clip,scale=0.3]{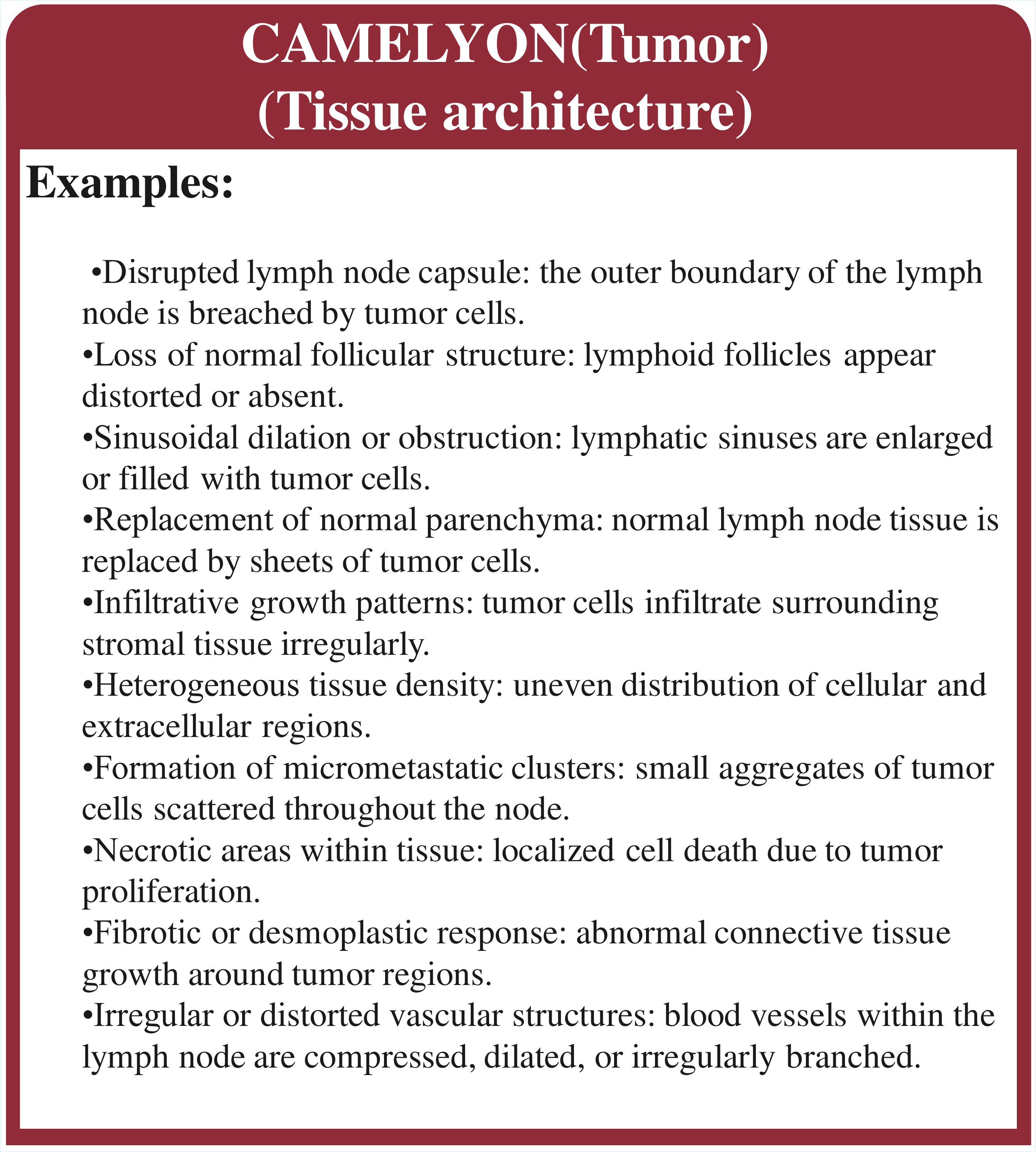} 
	\label{tumor2}
\end{figure*}

\begin{figure*} 
	\centering
	\includegraphics[trim=0 0 0 0,clip,scale=0.3]{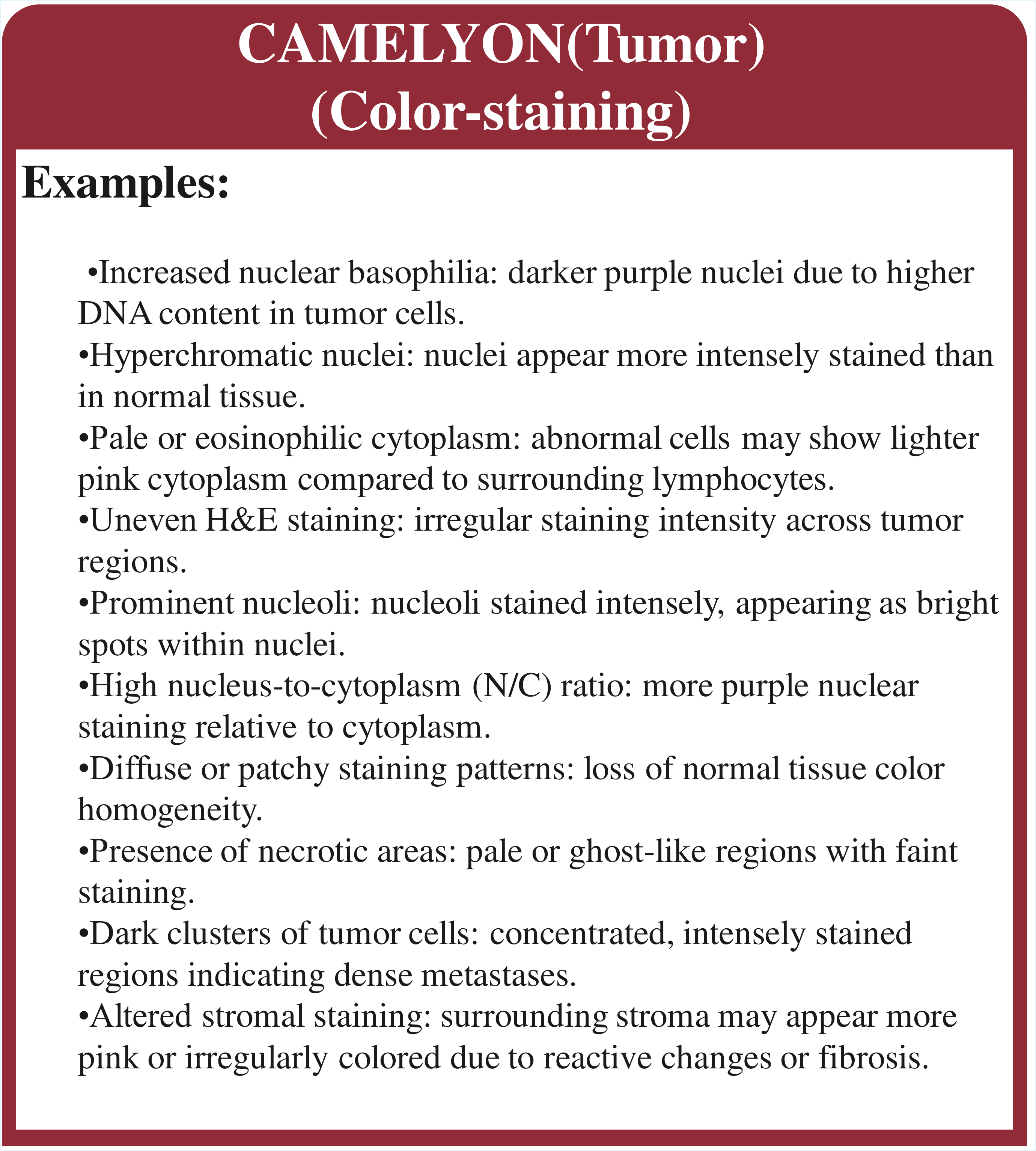} 
	\label{tumor3}
\end{figure*}

\begin{figure*} 
	\centering
	\includegraphics[trim=0 0 0 0,clip,scale=0.3]{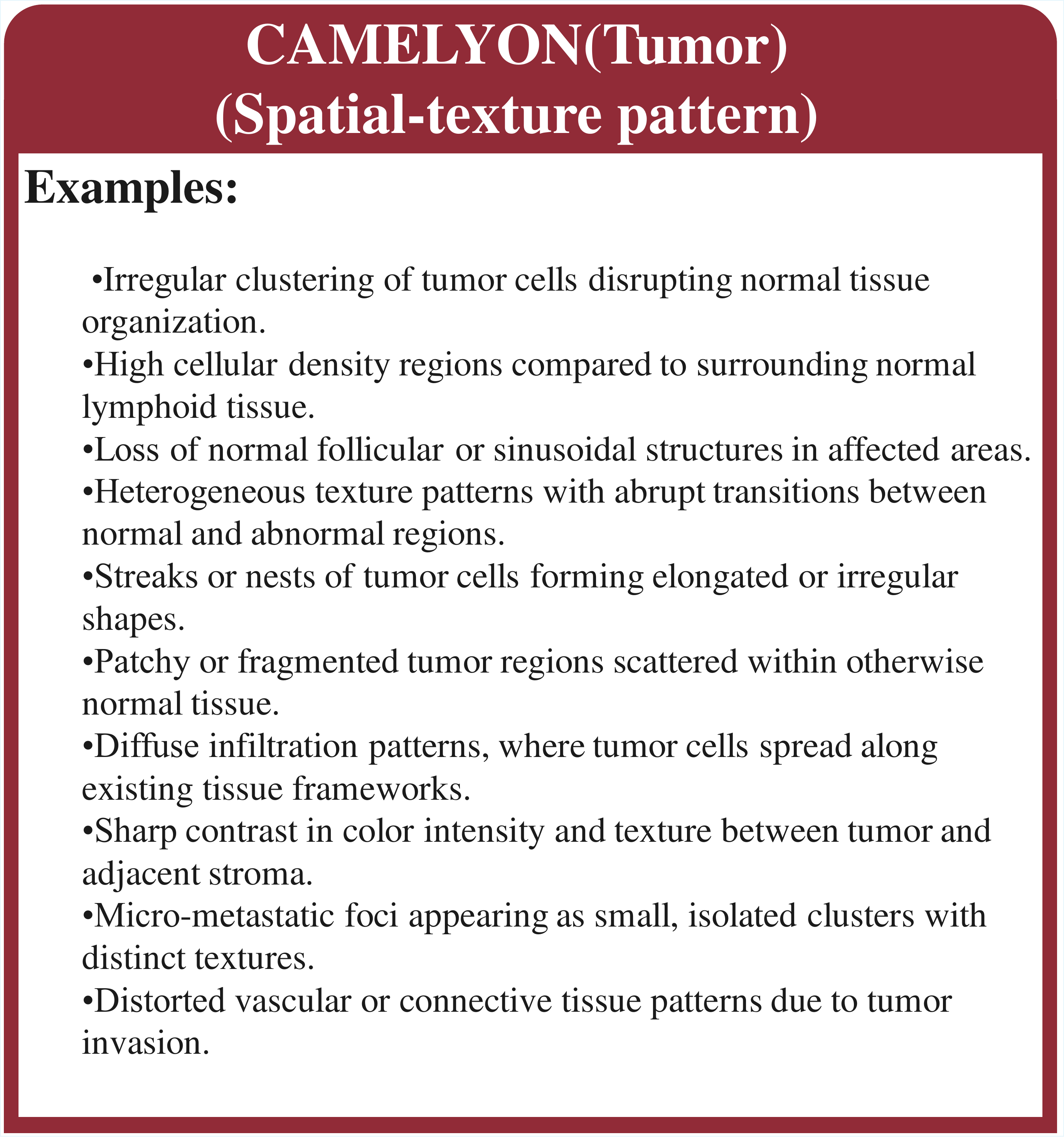} 
	\label{tumor4}
\end{figure*}